# Deep Learning in Renewable Energy Forecasting: A Cross-Dataset Evaluation of Temporal and Spatial Models

Lutfu S. Sua
*Dept. of Management and Marketing, Southern University and A&M College, Baton Rouge, LA, USA*
Haibo Wang
*Dept. of International Business and Technology Studies, Texas A&M International University, Laredo, Texas, USA*
Jun Huang
*Dept. of Management and Marketing, Angelo State University, San Angelo, TX, USA*

**Abstract**

Unpredictability of renewable energy sources coupled with the complexity of those methods used for various purposes in this area calls for the development of robust methods such as DL models within the renewable energy domain. Given the nonlinear relationships among variables in renewable energy datasets, DL models are preferred over traditional machine learning (ML) models because they can effectively capture and model complex interactions between variables. This research aims to identify the factors responsible for the accuracy of DL techniques, such as sampling, stationarity, linearity, and hyperparameter optimization for different algorithms. The proposed DL framework compares various methods and alternative training/test ratios. Seven ML methods, such as Long-Short Term Memory (LSTM), Stacked LSTM, Convolutional Neural Network (CNN), CNN-LSTM, Deep Neural Network (DNN), Multilayer Perceptron (MLP), and Encoder-Decoder (ED), were evaluated on two different datasets. The first dataset contains the weather and power generation data. It encompasses two distinct datasets, hourly energy demand data and hourly weather data in Spain, while the second dataset includes power output generated by the photovoltaic panels at 12 locations. This study deploys regularization approaches, including early stopping, neuron dropping, and *L2* regularization, to reduce the overfitting problem associated with DL models. The LSTM and MLP models show superior performance. Their validation data exhibit exceptionally low root mean square error values.

**Keywords:** deep learning (DL); renewable energy; Long-Short Term Memory (LSTM); Convolutional Neural Network (CNN); Deep Neural Network (DNN); Multilayer Perceptron (MLP); Encoder-Decoder (ED).

1. Introduction

Successful integration of renewable energy sources to the electrical grid requires accurate reliability and cost predictions. However, due to the irregular nature of these sources, forecasting energy output poses challenges. Nevertheless, this is a vital task in order to preserve grid stability, minimize the greenhouse effect, and maximize the effectiveness of energy management (Putz et al., 2023). To this end, methods such as time series analysis, machine learning (ML), deep learning (DL), and neural networks (NN) are being used to overcome this challenge. In the context of renewable energy forecasting, pertinent characteristics include:

- Time Series Data: historical energy output, time of day, day of week, and season
- Spatial Data: location, altitude, and distance to the nearest weather station
- Meteorology Data: temperature, solar radiation, humidity, and wind speed
- Derived Features: derivatives, integrals, and statistical metrics

While the advantages of DL are evident, challenges such as data scarcity, model interpretability, and computational demands persist. Overcoming these challenges requires hybrid techniques and open access to high-quality datasets as well as advanced model architecture.

The primary objective of this study is to develop accurate and reliable prediction models that can contribute to a more sustainable energy future by improving energy management and minimizing greenhouse gas emissions. To help overcome this challenge, by using DL methods, we propose the following research question (RQ):

*RQ: What is the performance of DL algorithms in predicting power output with spatial data?*

Aligned with this objective, this research proposes an analytics framework to integrate a number of basic DL algorithms with and without regularization approaches to determine critical factors pertaining to the reliability and availability of renewable energy output forecasts. This framework combines DL with sampling techniques to mitigate methodology-driven bias, a standard limitation in existing algorithms. To address the overfitting problem associated with DL models, this study deploys regularization approaches and analyzes the tradeoff between overfitting and accuracy. Using DL, the framework can effectively capture non-linear relationships between energy consumption and various factors, including weather, seasonality, and usage patterns, common in energy data.

The rest of the paper is organized in the following way: Section 2 provides theoretical background, while Section 3 presents information about DL methods. Section 4 compares various DL methods through experimental analyses. Finally, Section 5 presents the conclusions and potential directions for future research.

## 2. Theoretical Background

ML methods have successfully been employed in energy systems' planning, reliability, and security. This study employs meta-learning and DL combinations in a multivariate time series predicting renewable energy demand and supply. DL methods offer great potential as a solution to such problems due to their ability to learn behavioral patterns and adapt by detecting anomalies existing in the system. Behrens et al. (2024) proposed incorporating endogenous technological learning into energy system modeling, focusing on renewable energy areas, analyzing the methods for representing cost reductions due to learning-by-researching and learning-by-doing, which introduce non-linearities into models, making optimization difficult. Spatial aggregation and decomposition methods are suggested to maintain computational feasibility. The process by which an organism develops a mental image of its surroundings is known as spatial learning. Both invertebrate and vertebrate species have been shown to exhibit spatial learning (Brodbeck, 2012).

DL, a subset of ML and artificial intelligence, has become an important technology due to its accuracy and adaptability by leveraging multi-layered NN to model complex patterns and relationships in data. Surveys on ML methodologies reveal its versatility and growing adoption in energy research (Ying et al., 2023; Aslam et al., 2021). Their work sheds light on the advantages of NN in capturing nonlinear patterns and adapting to evolving datasets. Renewable energy systems, particularly solar and wind, are subject to variability and require accurate forecasting models to manage variability. Dairi et al. (2020) present a short-term forecasting of solar power production by using a Variational AutoEncoder, which is a deep learning method providing flexible nonlinear approximation. Miroshnyk et al. (2022) developed NN-based models to predict energy outputs over short timescales, enabling more reliable grid integration. Hybrid models such as CNN-LSTM frameworks have also been employed, as seen in Agga et al. (2022), for photovoltaic power forecasting, showcasing their ability to capture spatiotemporal dependencies in energy data. Abdelkader et al. (2024) apply three GNN models, Graph Convolutional Networks (GCN), Graph Attention Networks (GAT), and GraphSage, for spatiotemporal photovoltaic energy prediction. Chang et al. (2021) developed DL models to predict solar power generation, underscoring the significance of large-scale data processing capabilities and the role of DL in making accurate predictions under variable weather conditions. Nam et al. (2020) also proposed a DL framework to predict renewable energy outputs. Their results confirmed that the framework can significantly enhance the reliability of renewable energy forecasting, providing a stable foundation for energy planning and decision-making. Zhang et al. (2018) demonstrated how convolutional and recurrent NN can effectively predict solar energy output through weather and historical data to train models. The reported findings suggest that these models can outperform traditional statistical methods in computational efficiency and accuracy. Comparative analyses of forecasting techniques are vital for evaluating the efficacy of different algorithms. Quantile estimation of renewable energy production has also been explored, with Alcantara et al. (2023) utilizing deep neural networks (DNN) to predict regional outputs. These studies highlight the importance of model choice and customization based on specific energy systems and datasets.

While DL has succeeded remarkably, challenges like computational requirements, data quality, and model interpretability still exist. Phan et al. (2022) proposed novel frameworks incorporating data preprocessing and postprocessing techniques to address these issues. Table 1 provides a comparison summarizing previous approaches, their strengths, limitations, and how the proposed method improves upon them.

Table 1. Comparison of previous work

| Study / Model | Strengths | Limitations | Improvement |
| --- | --- | --- | --- |
| LSTM (Kong et al., 2019) | Effective for long-term dependency modeling; widely used for time-series data | Struggles with high-dimensional input and overfitting on limited datasets | Applies regularization and dataset ratio testing to prevent overfitting and validate generalizability |
| Stacked LSTM (Yu et al., 2019) | Deeper learning capacity, better abstraction | Requires more training data, risk of overfitting | Demonstrates consistent low RMSE across varying test/train splits; optimal in lower ratio scenarios |
| CNN (Abdoos et al., 2024) | Captures spatial hierarchies, fast computation | Not ideal for sequential forecasting on its own | Integrated into a hybrid CNN-LSTM model for spatiotemporal learning |
| CNN-LSTM (Agga et al., 2022) | Combines spatial and temporal features effectively | Model complexity can increase; tuning is required | Improved prediction at smaller training ratios; stable across datasets |
| Encoder-Decoder (Wang et al., 2022) | Good for sequence-to-sequence modeling; handles variable input/output lengths | Computationally intensive; overfitting with small data | Best performer at higher ratios; regularization enhances generalizability |
| DNN (Oluleye et al., 2023) | Good baseline for nonlinear relationships | Weak on sequence learning; high overfitting risk | Regularization significantly improves RMSE at higher ratios |

| Time-Distributed MLP (Chan et al., 2023) | Simple, robust in spatially distributed settings | Poor temporal awareness; limited for long sequences | With regularization, it achieves the best RMSE on a photovoltaic dataset (Dataset-2) |
| --- | --- | --- | --- |
| Phan et al. (2022) | Integrated pre/postprocessing to improve solar forecasting accuracy | Not fully automated or generalized across datasets | The proposed model pipeline is fully automated, scalable across datasets with PCA, mutual info, and stationarity analysis |
| Alcantara et al. (2023) | Focuses on quantile prediction for uncertainty analysis | Less suitable for deterministic forecasting | The current model emphasizes deterministic RMSE reduction while being adaptable for uncertainty modeling |
| Chang et al. (2021) | Demonstrates DL capability for large-scale solar data | Lacks flexible ratio testing or regularization benchmarks | Introduces controlled test/train variations and regularized tuning for realistic deployment scenarios |

### 2.1 Climate Factors

The reliance of renewable energy generation on weather conditions increases the complexity of the energy supply and demand balance. Forecasting models must consider climatic factors playing a role in the energy supply. As such, the solar panel capacity is determined not only by the temperature, but also by the sunlight they receive, as well as factors such as barometric pressure and humidity (Jathar et al., 2023). Predicting wind energy output is one of the biggest challenges that constitutes a barrier to investment in this area. Thus, accurate forecasting is essential to foresee the energy demand when the power generated by the wind turbines is not expected to meet the demand.

### 2.2 Renewable Energy Production

DL methods are applied in the renewable energy industry in areas such as the design of the infrastructure based on forecasting energy demand and weather conditions, anomaly detection and failure prediction in energy systems, production forecasting by utilizing satellite data, and distribution network optimization using production and consumption data (Liu et al., 2020), and expected market prices (Wang et al., 2022).

Nearly limitless sources of solar energy, coupled with the continuously decreasing cost of panel installation due to improved research and development, are among many reasons solar energy has been among the top three energy sources added to the grid in recent years (EERE, 2020). Thus, solar power applications' global capacity supports the energy supply and meets the labor market for sufficient development (Maka and Alabid, 2022). Meanwhile, in order to balance supply and demand, manage grid stability, and optimize energy dispatch, grid operators need reliable forecasts (Achouri et al., 2023). However, solar output is predicted to increase with the intensifying reliance on this energy source. With the uncertain nature of wind power and speed, DL applications in the industry tend to focus on increasing reliability by predicting wind behavior (Zhang et al., 2021).

Chen et al. (2020) utilized a deep architecture with a multi-task and multi-model learning approach to forecast wind power. Du (2019) used the weather prediction model and meteorological data to predict wind power. Khodayar and Wang (2019) proposed a graph DL model for the probabilistic behavior of the wind speed. Schlechtingen et al. (2013) evaluated several data-mining methods for wind turbine power curve monitoring, to determine the power output. Chen et al. (2021) developed another DL technique combining spatiotemporal correlation and using a two-stage modeling approach to forecast multiple wind turbines.

Another area where DL is utilized is the prediction of tidal power, which offers a viable alternative and reliable source for the growing energy needs. As such, based on stream regimes, the DNN approach was used to forecast the design values of tidal power plants (Fujiwara, 2022). As a result of the unpredictable nature of hybrid renewable energy systems, they provide greater hurdles (Thirunavukkarasu et al., 2023). These difficulties result in the growing interest in DL applications. Bansal (2022) proposed a hybrid system based on photovoltaics and wind for system optimization. Shakibi et al. (2023) utilized an ANN method for optimizing a solar and wind plant for hydrogen production. Behzadi et al. (2023) developed an intelligent building system with low-temperature heating and high-temperature cooling to utilize renewable energy. Yao et al. (2023) introduced main performance indicators to compare ML workflows for energy research and evaluated their application in harvesting, storing, and converting energy. Bilgic et al. (2023) reviewed ANN advances in hydrogen production research.

### *2.3 Steps of Automated DL Models*

This section introduces the DL techniques employed in the analysis. DL applications in most domains require automated DL techniques to ease complex and lengthy processes. As part of the first step, training and testing sets are created for model validation. This way, DL methods can be trained on a training set, and their accuracy can be assessed by a testing set. Keeping a validation dataset prevents overfitting of the training set and aids in evaluating the models' prediction ability (Joseph, 2022). Two datasets utilized here are divided into training and test sets, investigating each data split configuration's impact on the DL techniques. To ensure the robustness and generalizability of DL models, model performance is evaluated across multiple training/test split ratios: 0.2, 0.3, 0.4, and 0.5, which correspond to using 20%, 30%, 40%, and 50% of the data, respectively, for validation. These ratios are selected based on a combination of standard practices in the time-series forecasting literature and the ample size of both datasets. This enables meaningful partitioning without compromising model stability. Lower ratios are particularly informative in assessing the ability of a model to generalize from limited data. Conversely, higher ratios simulate conditions with abundant training data, helping evaluate the scalability and performance ceiling of each model. This approach allows a detailed understanding of model behavior under varying data availability situations, which is a key concern in renewable energy forecasting.

Categorical data must be encoded into numerical values before utilizing the model. In a labeled dataset, the supervised encoding approach utilizes the provided labels to transform categorical variables into numerical representations. Leave-one-out encoders are utilized for the datasets in this study.

Feature selection determines the most important factors existing in the dataset. This step influences the model performance by limiting the data dimension while determining the optimal energy features. Feature selection involves selecting only those that influence power production. Thus, a correlation matrix is developed to analyze how some features can explain others and to determine relationships. Moreover, weather sensor data and solar power generation data are merged. A factor is removed from the correlation matrix if it is not correlated with all the features.

### 3. Research Design

This study introduces a new automated model development pipeline framework, including regularization and embedding. Figure 1 illustrates such a flowchart. This framework streamlines data processing, model construction, and inference deployment. To support the reproducibility of the proposed deep learning framework, comprehensive data preprocessing procedures were systematically implemented. These steps ensured consistent model performance across datasets and provided a robust foundation for temporal and spatial analysis. Data normalization was conducted using Min-Max scaling, transforming each feature into the [0,1] interval. This technique was selected due to its compatibility with activation functions commonly used in deep learning architectures, and its effectiveness in accelerating convergence during training. Importantly, normalization parameters were computed exclusively from the training set and applied independently to the validation and test sets to mitigate data leakage. Target variables were also normalized prior to training and subsequently denormalized after inference to facilitate evaluation in original units.

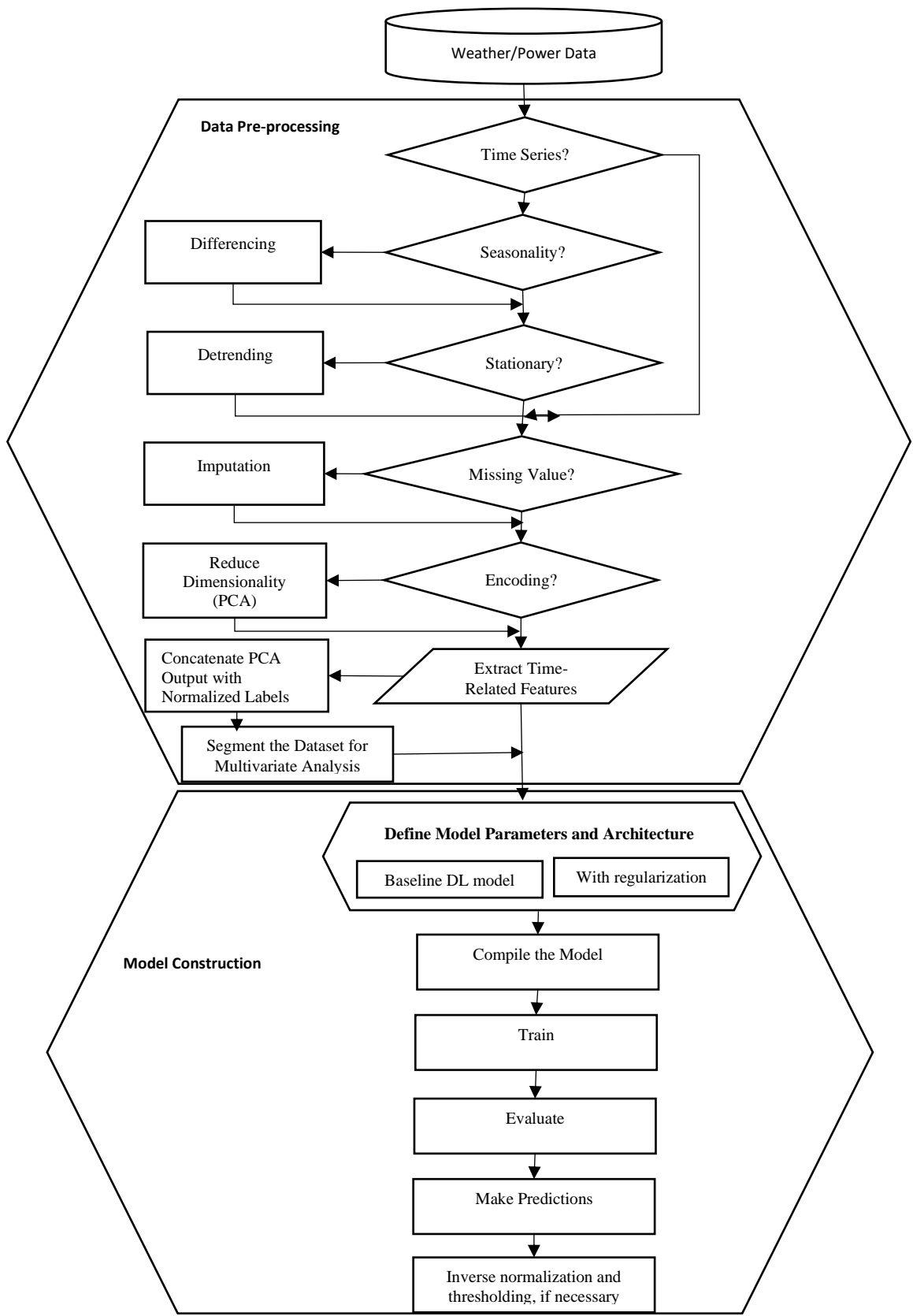

Figure 1. Statistical and DL Framework

Handling missing data is essential for the reliability of time-series models. A hybrid imputation strategy was employed. Forward-fill interpolation was used for time gaps comprising up to three consecutive missing timestamps, preserving local temporal continuity. For more extensive missing segments, mean imputation and mode imputation were applied for continuous variables and categorical features, respectively. Features with more than 15% missing values were excluded from further analysis to maintain data integrity. A range of feature engineering methods was utilized to enhance the model's ability to capture temporal, cyclical, and spatial patterns:

- Temporal Features: Trigonometric transformations (*sine_hr, cos_hr, sine_mon, cos_mon*) were derived from the timestamp to encode periodic variations associated with daily and monthly cycles. Binary flags indicating weekdays, weekends, and meteorological seasons were also created.
- Spatial and Categorical Encoding: Location-based categorical variables were converted into numerical values using leave-one-out encoding, a supervised encoding technique that mitigates overfitting and data leakage while preserving predictive relevance.
- Stationarity Adjustments: Based on the results of KPSS and ADF tests, differencing and detrending techniques were selectively applied to achieve stationarity in non-stationary time-series features.
- Dimensionality Reduction: Principal Component Analysis (PCA) was applied to reduce the original 58-feature set into 13 orthogonal components that explained approximately 80% of the variance. This step enhanced computational efficiency and minimized multicollinearity.
- Feature Importance Assessment: A correlation matrix was generated to identify and remove redundant features with low correlation coefficients ($< 0.1$) relative to the target variable. Additionally, mutual information analysis was performed to quantify the dependence between climate variables and energy outputs, further informing the selection of high-impact predictors.

These preprocessing protocols formed a critical component of the automated model development pipeline and were uniformly applied across both datasets to ensure consistency, reduce overfitting, and improve generalization capability of the DL models.

### 3.1 Data Sources

The first dataset, containing the weather and power generation data, encompasses two distinct datasets, hourly energy demand data and hourly weather data, spanning four years, from 2015 to 2018, in Spain[i]. Hourly energy demand data includes information on electricity obtained from various renewable sources, total energy load, price of electricity, etc., while weather data contains conditions such as temperature, rainfall, and humidity for five Spaniard cities, including Madrid, Valencia, Barcelona, Bilbao, and Seville (Sartini, 2024). Tables 2 and 3 provide the descriptive weather and power generation statistics, respectively. After merging and preprocessing these datasets, the resulting dataset with 58 features and 35,064 rows was used to predict power consumption. Seven DL methods were evaluated on this dataset.

Table 2. Descriptive Statistics of Dataset 1 for Weather

| Weather | Temp | Humidity | Wind_Speed | Wind_Deg | Rain_One hr | Rain_Three hr | Snow_Three hr | Clouds_all |
|---|---|---|---|---|---|---|---|---|
| Mean | 289.6000 | 68.4200 | 2.4700 | 166.5 | 0.0754 | 0.0003 | 0.0047 | 25.0730 |
| Median | 289.15 | 72 | 2 | 177 | 0 | 0 | 0 | 20 |
| Std. | 8.0260 | 21.9000 | 2.0950 | 116.6 | 0.3988 | 0.0072 | 0.2226 | 30.7740 |
| Skewness | 0.2250 | -0.5254 | 3.1704 | -0.0314 | 15.8932 | 184.1125 | 68.8350 | 0.9468 |
| Kurtosis | -0.3769 | -0.6173 | 97.1242 | -1.3661 | 399.4752 | 57067.4057 | 5470.8889 | -0.5790 |
| Min | 262.2400 | 0 | 0 | 0 | 0 | 0 | 0 | 0 |
| Max | 315.6000 | 100 | 133 | 360 | 12 | 2.315 | 21.50 | 100 |

Table 3. Descriptive Statistics of Dataset 1 for Power Generation

| Power Generation | Hydro | Solar | Wind | Other Renewable |
|---|---|---|---|---|
| Mean | 4,050 | 1,432.8 | 5,461.6 | 85.630 |
| Median | 3587 | 616 | 4847 | 88 |
| Std. | 2,115.4 | 1,679.9 | 3,215.2 | 14.077 |
| Skewness | 0.8098 | 1.0203 | 0.7835 | -0.2160 |
| Kurtosis | -0.0438 | -0.3877 | 0.0574 | -0.8266 |
| Min | 0 | 0 | 0 | 0 |
| Max | 11,613 | 5,792 | 17,436 | 119 |

The KPSS and ADF tests are utilized as stationarity tests. The null hypothesis in the linear regression-based KPSS test is that the time series is stationary (unlike ADF), meaning p-values lower than the significance level indicate a non-stationary series. The number of lag periods used in the test is important as it helps adjust the test to capture the time series dynamics better.

Generation Data (generation_hydro, generation_solar, etc.) shows that all variables related to renewable energy generation are stationary based on their very small p-values (Appendix, Table A1). Weather Data (temp_Barcelona, humidity_Seville, etc.) shows the same pattern where the p-values are all very small, meaning most weather-related variables are stationary. Variables like clouds_all_Barcelona, clouds_all_Bilbao, and clouds_all_Valencia, have very large negative ADF statistics, and the p-values are either 0 or close to it, showing that these variables are also stationary. Based on the p-values, all the time series tested (for both renewable energy generation and weather variables) appear stationary, making them suitable for time series modeling and forecasting techniques that assume stationarity.

The second dataset involves power output data from solar panels placed in 12 cities over 14 months[ii]. The power outcome of the panels, wind speed, date, season, time sampled, location, latitude, longitude, altitude, ambient temperature, humidity, visibility, pressure, and cloud ceiling are some of the independent variables of the dataset (Williams et al, 2019). This dataset has 17 features and 21,045 samples and is utilized in forecasting the photovoltaic panels' power output. Table 4 presents the descriptive statistics for numerical variables.

Table 4. Descriptive Statistics of Dataset 2 for Panels

| Variable | Power output (Watt) | Humidity (%) | Ambient temp (C) | Wind speed (km/h) | Visibility (km) | Pressure (millibar) | Cloud ceiling (km) |
|---|---|---|---|---|---|---|---|
| Mean | 12.9785 | 37.1219 | 29.2851 | 10.3183 | 9.7000 | 925.9447 | 515.9668 |
| Median | 13.7987 | 33.1237 | 30.2891 | 9 | 10 | 961.1 | 722 |
| Std. | 0.0491 | 0.1642 | 0.0852 | 0.0440 | 0.0093 | 0.5874 | 2.0811 |
| Skewness | -0.0353 | 0.6652 | -0.3264 | 0.6270 | -5.1447 | -0.3588 | -0.8224 |
| Kurtosis | -1.0822 | -0.2626 | 0.16133 | 0.5282 | 27.2766 | -1.5580 | -1.2527 |
| Minimum | 0.2573 | 0 | -19.9818 | 0 | 0 | 781.7 | 0 |
| Maximum | 34.2850 | 99.9877 | 65.7383 | 49 | 10 | 1029.5 | 722 |

ADF test results show that location variables generally have higher p-values, likely due to their categorical nature, which often does not adhere to stationarity tests like continuous variables (Appendix, Table A2). Trigonometrical variables (sine_hr, cos_hr, sine_mon, cos_mon) are stationary due to their periodic nature. Most of the weather, seasonal-related variables, and some periodic (sine/cosine) variables, are stationary. Variables such as Latitude, Pressure, and Location are likely non-stationary or do not meet the strict stationarity assumption for time series analysis. The graphic results of correlation among variables for dataset-1 and dataset-2 are provided in the Appendix, illustrated by Figures A5 and A6.

### 3.2 Prediction Models

Predictive modeling efficiency has significantly increased due to advances in computational power and data availability. Meanwhile, a classification or regression model may be a better fit, based on the nature of the dependent variable. The DL prediction models utilized in this research are for power generation and are introduced as follows:

In this study, we deployed seven DL models. Compared to traditional ML techniques, DL models can effectively handle non-linear relationships between variables, which are common in time series data. Additionally, DL models can achieve higher accuracy than traditional ML models, especially on complex datasets, and handle high-dimensional data with multiple variables, making them well-suited for multivariate time series analysis. Regularization techniques are utilized here to prevent overfitting in DL techniques. Poor performance on unseen data results from overfitting, which happens when a model becomes overly complicated and learns the noise in the training data. Regularization helps to prevent overfitting by introducing a penalty term to the loss function to avoid large weights. This approach improves the model's generalization to unseen data by reducing the impact of noise and outliers. Regularization also reduces the complexity of the model by eliminating unnecessary weights and connections, which in turn improves the interpretability of the model by reducing the number of features and weights. Moreover, regularization can help reduce the model's training time by reducing the number of parameters to be optimized. Dropout regularization is utilized in particular, randomly setting a fraction of the weights to zero during the DL training process to prevent overfitting. The DL prediction models used in this research for power generation are introduced in Appendix (Table A3).

RMSE and Loss values are two important metrics utilized to assess the performance of DL models in this study. RMSE measures the discrepancy between the expected and actual values. The square root of the mean of the squared discrepancies between the actual and expected values is how it is computed. Loss values measure the difference between the predicted and the actual values and are used to train DL models. The most common loss function used in DL is the Mean Squared Error (MSE) loss function. RMSE measures the difference between the predicted and actual values, while Loss values measure the difference between the predicted and actual values and are used to train DL models.

$$MSE = \sqrt{\frac{1}{n} * \Sigma(y_{true} - y_{pred})^2} \qquad (1)$$

where: $y_{true}$ is the actual value, $y_{pred}$ is the predicted value, n is the number of samples, $\Sigma$ denotes the sum of the squared differences. The most common loss function utilized in DL is the Mean Squared Error (MSE) loss function. RMSE is the square root of the Loss value.

$$MSE\ Loss = \frac{1}{n} * \Sigma(y_{true} - y_{pred})^2 \qquad (2)$$

### 3.3 ML Analysis of Climate Factors

Mutual information of a pair of random variables quantifies the amount of information related to one of them by tracking the other. It measures the mutual dependence of weather (W) and energy production (E). It helps determine how different the joint distribution of the (E, W) pair is from the product of their marginal distributions (Chidanand et al., 2021). If the joint distribution of (E, W) is P(E), and their marginal distributions are $P_{(E)}$ and $P_{(W)}$, the mutual information can be represented as:

$$(E; W) = D_{KL}(P_{(E,W)}) \| P_E \otimes P_W \tag{3}$$

where $D_{KL}$ is the Kullback-Leibler divergence. The mutual information of the jointly discrete random variables is calculated as a double sum:

$$I(E; W) = \sum_{w \in W} \sum_{e \in E} P_{(E,W)}(e, w) \log \log \left( \frac{P_{(E,W)}(e,w)}{P_{(E)} P_{(W)}(w)} \right) \tag{4}$$

where $P_{(E,W)}$ is the joint probability mass function, and $P_{(E)}$ and $P_{(W)}$ are the marginal probability mass functions of E and W, respectively. A double integral replaces the double sum:

$$I(E; W) = \int_w \int_e P_{(E,W)}(e, w) \log \log \left( \frac{P_{(E,W)}(e,w)}{P_{(E)} P_{(W)}(w)} \right) de dw \tag{5}$$

where $P_{(E,W)}$ is the joint probability density function, and $P_{(E)}$ and $P_{(W)}$ are the marginal probability density functions of E and W, respectively. The heatmaps in Figure A1 and Figure A2 (Appendix) illustrate the mutual information from Equations 3-5 for both datasets.

The paper uses feature importance analysis to some degree. It applies mutual information analysis (Equation 3–5) to identify dependencies between climate variables and energy outputs. It also generates correlation matrices to identify redundant features and improve feature selection. PCA is applied for dimensionality reduction, which helps indirectly interpret dominant components. While this study provides a comparative analysis of predictive performance across multiple deep learning models, the inherent "black-box" nature of these models remains a limitation for practical deployment in energy systems. Although mutual information and correlation matrices were employed for initial feature selection, the models themselves lack post-hoc interpretability mechanisms. Future research should incorporate explainability techniques such as SHAP (SHapley Additive exPlanations) values or LIME (Local Interpretable Model-agnostic Explanations) to identify which features most significantly influence predictions. Furthermore, integrating attention mechanisms within recurrent or hybrid architectures (e.g., Attention-LSTM or Transformer-based models) can improve both model transparency and performance by highlighting relevant temporal or spatial patterns. These tools can enhance stakeholder trust and facilitate data-driven decision-making in renewable energy planning and management. Principal Component Analysis (PCA) reduces the dimensionality of a dataset with many features, simplifying the complexity of high-dimensional data while retaining patterns and trends (Jeon et al., 2022). This approach transforms the original 58 features into 13 uncorrelated components, preserving 80% of the variability in the data. The synthesized components are consequently used in the model. Figures A3 and A4 (Appendix) provide different ratios' results for both datasets.

### 3.4 Hyperparameter Optimization Strategy

The performance of deep learning models is highly sensitive to the configuration of their hyperparameters. To systematically optimize these parameters and avoid suboptimal performance, a two-stage tuning strategy was employed. Initially, a random search method was utilized to broadly explore the hyperparameter space due to its efficiency in identifying promising regions compared to exhaustive methods such as grid search. This stage included ranges for key parameters such as the number of layers (2–5), number of neurons per layer (32–256), learning rate (0.0001–0.01), dropout rate (0.1–0.5), batch size (32, 64, 128), and activation functions (ReLU, tanh, sigmoid).

Subsequently, a grid search refinement was performed within the most promising parameter ranges identified during random search to fine-tune model performance. This hybrid approach balanced computational feasibility with tuning precision. All model evaluations during hyperparameter tuning were based on validation RMSE, using stratified cross-validation across multiple training/test splits to ensure stability. The final hyperparameter configurations for each model were selected based on the lowest average validation RMSE while also considering overfitting risk, training time, and generalization performance. Table A4 provides the hyperparameter configurations for DL models used in this study.

### 4. Managerial Implications and Discussion

Clean energy demand is expected to increase in parallel with the decrease in investment costs resulting from government policies and technological advances. The renewable energy market is forecasted to reach $1.98 trillion by 2030 (Deloitte, 2023). The weather model of the National Oceanic & Atmospheric Administration is expected to reach $150 million in annual energy savings (Jeon et al., 2022). Meanwhile, these investments can be limited by potential barriers related to the political and financial uncertainties, causing supply chain disruptions on a global scale. These barriers include, but are not limited to, high interest and inflation rates, tightened trade policies, and delayed energy projects. In such environments, predicting the patterns related to renewable energy becomes one of the most important components in feasibility studies while dealing with those barriers.

#### 4.1 Dataset-1

In addition to measuring the accuracy, Table 5 compares the RMSE values obtained from seven models at four different training/test ratios. To assess the statistical reliability of the reported RMSE values, 95% confidence intervals were computed based on repeated runs using five-fold cross-validation. For each model and training/test ratio, the mean and standard deviation of RMSE were calculated across five independent splits, and confidence intervals were derived using the formula:

$$CI = \bar{x} \pm 1.96 \, x \, \frac{\sigma}{\sqrt{k}} \tag{6}$$

Table 5. RMSE Values for Dataset-1

| Model | Ratio | Train RMSE (±95% CI) | Validation RMSE (±95% CI) |
|---|---|---|---|
| LSTM | 0.2 | 0.0439 ± 0.0018 | 0.0423 ± 0.0015 |
|  | 0.3 | 0.0537 ± 0.0021 | 0.0458 ± 0.0018 |
|  | 0.4 | 0.0474 ± 0.0019 | 0.0472 ± 0.0019 |
|  | 0.5 | 0.0516 ± 0.0020 | 0.0542 ± 0.0023 |
| Stacked LSTM | 0.2 | 0.0375 ± 0.0016 | 0.0388 ± 0.0014 |
|  | 0.3 | 0.0416 ± 0.0018 | 0.0376 ± 0.0015 |
|  | 0.4 | 0.0427 ± 0.0019 | 0.0438 ± 0.0018 |
|  | 0.5 | 0.0427 ± 0.0017 | 0.0488 ± 0.0020 |
| CNN-LSTM | 0.2 | 0.0363 ± 0.0015 | 0.0398 ± 0.0014 |
|  | 0.3 | 0.0431 ± 0.0017 | 0.0395 ± 0.0015 |
|  | 0.4 | 0.0423 ± 0.0018 | 0.0456 ± 0.0019 |
|  | 0.5 | 0.0451 ± 0.0020 | 0.0541 ± 0.0023 |
| Encoder-Decoder | 0.2 | 0.0354 ± 0.0014 | 0.0395 ± 0.0013 |
|  | 0.3 | 0.0396 ± 0.0016 | 0.0386 ± 0.0014 |
|  | 0.4 | 0.0382 ± 0.0015 | 0.0434 ± 0.0018 |

| | 0.5 | 0.0394 ± 0.0016 | 0.0489 ± 0.0021 |
|---|---|---|---|
| DNN | 0.2 | 0.0446 ± 0.0017 | 0.0427 ± 0.0016 |
| | 0.3 | 0.0510 ± 0.0019 | 0.0433 ± 0.0017 |
| | 0.4 | 0.0495 ± 0.0019 | 0.0500 ± 0.0021 |
| | 0.5 | 0.0497 ± 0.0021 | 0.0744 ± 0.0035 |
| Time-Distributed MLP | 0.2 | 0.0353 ± 0.0015 | 0.0411 ± 0.0015 |
| | 0.3 | 0.0369 ± 0.0016 | 0.0400 ± 0.0015 |
| | 0.4 | 0.0377 ± 0.0017 | 0.0443 ± 0.0018 |
| | 0.5 | 0.0337 ± 0.0016 | 0.0503 ± 0.0022 |
| ARIMA | 0.2 | 0.0325 ± 0.0014 | 0.1643 ± 0.0126 |
| | 0.3 | 0.0344 ± 0.0015 | 0.1543 ± 0.0112 |
| | 0.4 | 0.0363 ± 0.0016 | 0.1914 ± 0.0135 |
| | 0.5 | 0.0362 ± 0.0016 | 0.1594 ± 0.0121 |

The table indicates that Stacked LSTM consistently shows low val_rmse and val_loss across all ratios. For example, at a ratio of 0.2, it has a val_rmse of 0.0388 and 0.0376 at a ratio of 0.3, among the lowest. Another strong performer is the Encoder-Decoder, especially at a ratio of 0.4 (val_rmse = 0.0434) and a ratio of 0.5 (val_rmse = 0.0489), with minimal overfitting. CNN-LSTM performs well at smaller ratios, such as 0.2 (val_rmse = 0.0398) and 0.3((val_rmse = 0.0395), but slightly degrades at higher ratios. Another important observation is that regularization reduces overfitting but sometimes increases val_rmse—for example, Reg. CNN shows much higher train_rmse and val_rmse compared to CNN at most ratios. Smaller ratios (0.2, 0.3) generally lead to lower loss and RMSE due to a larger training data sample size. As the ratio increases, models tend to be slightly overfit (higher train_loss and val_loss differences). Stacked LSTM consistently shows good performance with low val_rmse (e.g., 0.0374 at 0.2 ratio and 0.0433 at 0.4 ratio), while DNN generally shows higher train_loss and val_loss, indicating it might not be the best choice for this task.

Based on the results, CNN-LSTM, Encoder-Decoder, and Stacked LSTM are the most effective architectures across different ratios, balancing low val_rmse and val_loss with minimal overfitting. Regularized versions can help reduce overfitting but may introduce additional errors if over-applied. Table 6 lists the best models for each ratio based on their val_rmse and val_loss values (smallest values preferred) while considering generalization, which requires smaller gaps between training and validation metrics.

Table 6. Summary of RMSE Results

| Ratio | Best Model | Val_RMSE |
|---|---|---|
| 0.2 | Stacked LSTM | 0.0388 |
| 0.3 | Stacked LSTM | 0.0376 |
| 0.4 | Encoder-Decoder | 0.0434 |
| 0.5 | Regularized DNN | 0.0472 |

Stacked LSTM has the lowest val_rmse value for ratio = 0.2 with a balanced train_rmse score of 0.0375 and good loss values. For ratio = 0.3, Stacked LSTM is the best model with low val_rmse and val_loss values, indicating excellent generalization. The results suggest that the Encoder-Decoder is a better model for higher ratios. It has the lowest val_loss with good generalization and slightly better performance for ratio = 0.4. The method also maintains strong performance for ratio = 0.5, with low val_rmse and minimal overfitting. DNN and Regularized DNN perform well for ratio=0.5, Figures 2-5 depict the prediction performance of the best models for each ratio provided in Table 5. The figures clearly show the exceptional performance of the models.

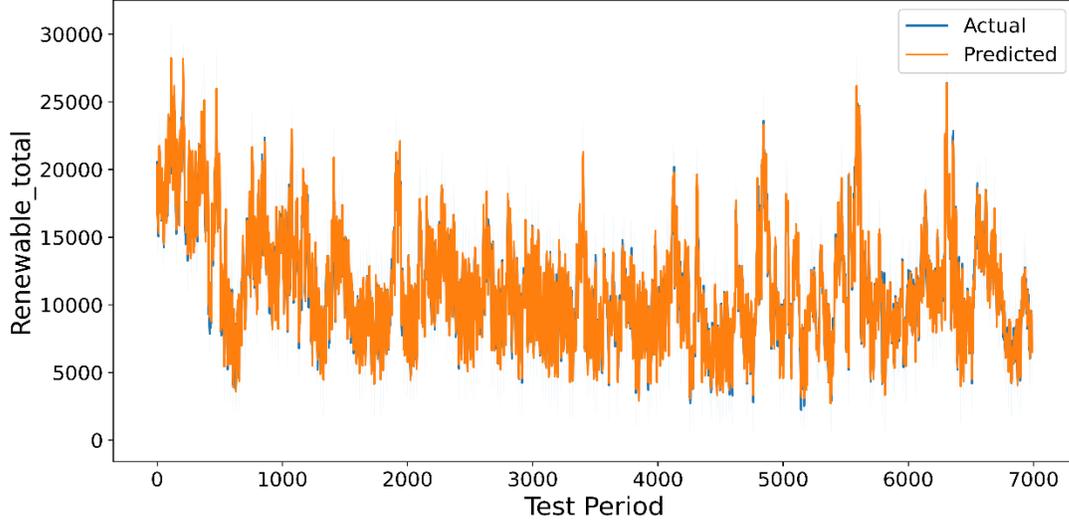

Figure 2. Stacked LSTM at 20% for Test Data

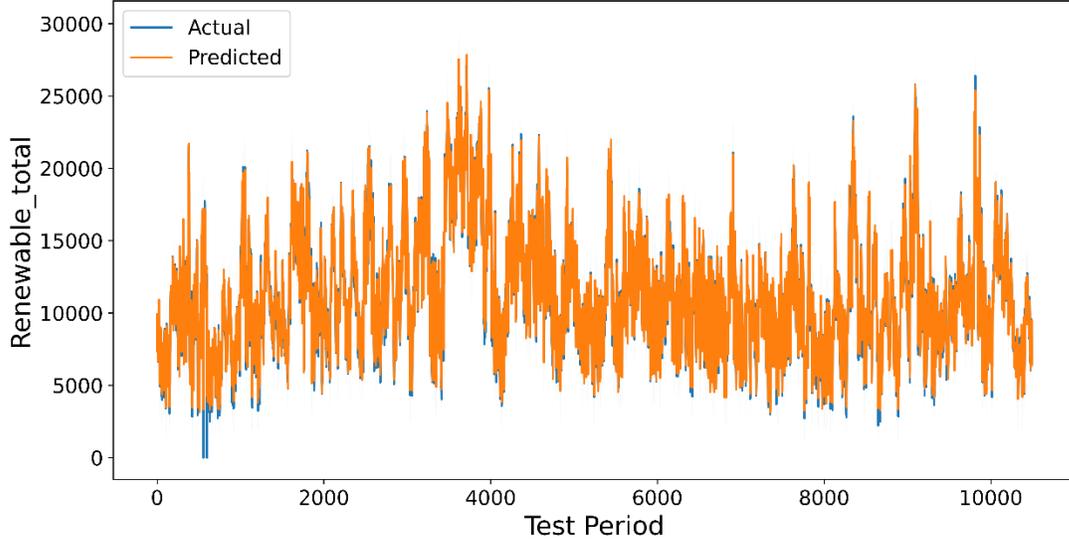

Figure 3. Stacked LSTM at 30% for Test Data

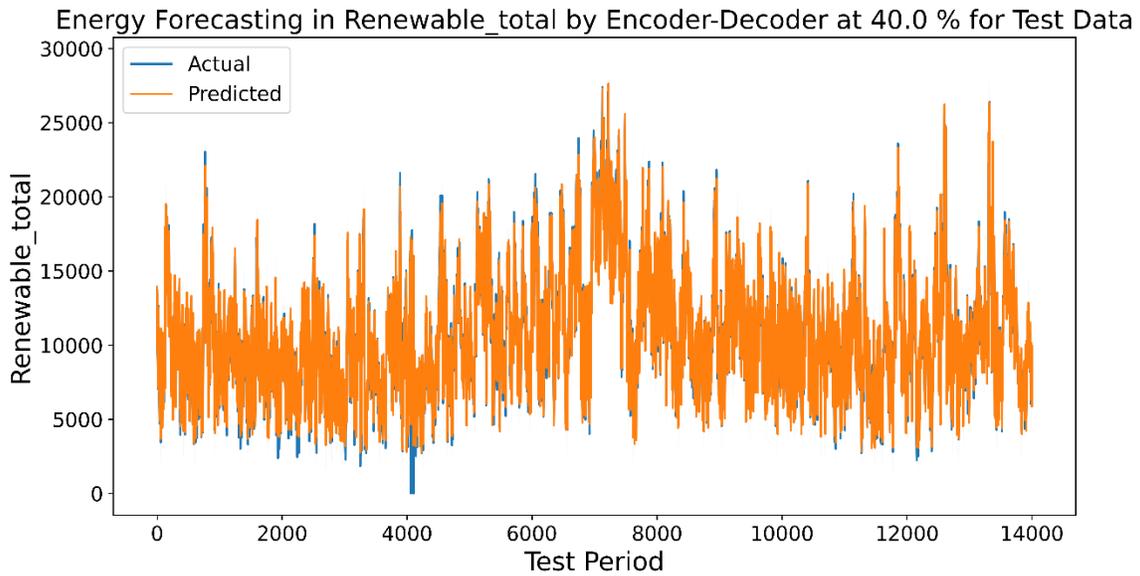

Figure 4. Encoder-Decoder at 40% for Test Data

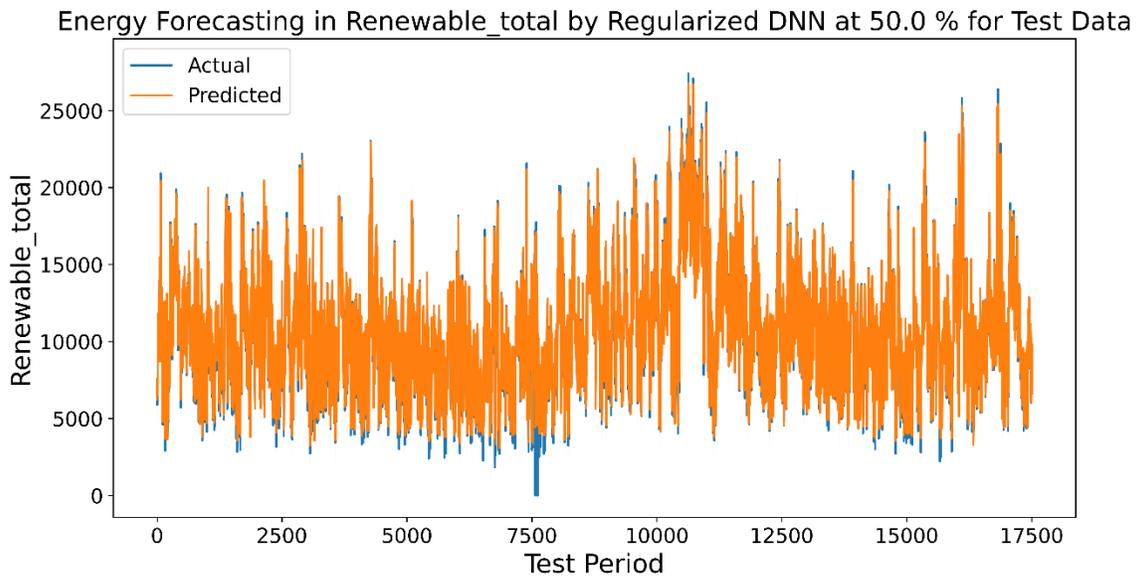

Figure 5. Regularized DNN at 50% for Test Data

### 4.2 Dataset-2

Applying DL techniques to the power outcome of the photovoltaic panels at twelve locations for power output forecast results in similar accuracy and test ratios. While the LSTM method performs well, the MLP method provides comparatively better results in almost all cases. Table 7 compares the RMSE values for the combined methods and training/test ratios.

Table 7. RMSE Values for Dataset-2

| Model | Ratio | Train RMSE (±95% CI) | Validation RMSE (±95% CI) |
| --- | --- | --- | --- |
| LSTM | 0.2 | 0.1519 ± 0.0045 | 0.1250 ± 0.0038 |
| | 0.3 | 0.1599 ± 0.0048 | 0.1454 ± 0.0043 |
| | 0.4 | 0.1464 ± 0.0042 | 0.1490 ± 0.0044 |

| | 0.5 | 0.1490 ± 0.0045 | 0.1698 ± 0.0050 |
| --- | --- | --- | --- |
| Time-Distributed MLP | 0.2 | 0.1484 ± 0.0043 | 0.1215 ± 0.0036 |
| | 0.3 | 0.1548 ± 0.0046 | 0.1387 ± 0.0042 |
| | 0.4 | 0.1491 ± 0.0044 | 0.1443 ± 0.0043 |
| | 0.5 | 0.1456 ± 0.0042 | 0.1755 ± 0.0052 |
| Reg. Time-Dist. MLP | 0.2 | 0.1539 ± 0.0045 | 0.1213 ± 0.0035 |
| | 0.3 | 0.1638 ± 0.0049 | 0.1405 ± 0.0041 |
| | 0.4 | 0.1534 ± 0.0045 | 0.1442 ± 0.0040 |
| | 0.5 | 0.1511 ± 0.0044 | 0.1702 ± 0.0050 |
| CNN-LSTM | 0.2 | 0.1478 ± 0.0042 | 0.1236 ± 0.0037 |
| | 0.3 | 0.1556 ± 0.0046 | 0.1476 ± 0.0044 |
| | 0.4 | 0.1495 ± 0.0043 | 0.1484 ± 0.0043 |
| | 0.5 | 0.1508 ± 0.0045 | 0.1743 ± 0.0051 |
| DNN | 0.2 | 0.1412 ± 0.0040 | 0.1249 ± 0.0036 |
| | 0.3 | 0.1428 ± 0.0041 | 0.1458 ± 0.0041 |
| | 0.4 | 0.1427 ± 0.0042 | 0.1480 ± 0.0042 |
| | 0.5 | 0.1357 ± 0.0041 | 0.1724 ± 0.0049 |
| Encoder-Decoder | 0.2 | 0.1526 ± 0.0044 | 0.1231 ± 0.0037 |
| | 0.3 | 0.1657 ± 0.0049 | 0.1430 ± 0.0042 |
| | 0.4 | 0.1524 ± 0.0043 | 0.1476 ± 0.0043 |
| | 0.5 | 0.1532 ± 0.0045 | 0.1703 ± 0.0048 |
| ARIMA | 0.2 | 0.1343 ± 0.1343 | 0.1439 ± 0.0047 |
| | 0.3 | 0.1372 ± 0.0037 | 0.1417 ± 0.0045 |
| | 0.4 | 0.1380 ± 0.0038 | 0.1401 ± 0.0044 |
| | 0.5 | 0.1400 ± 0.0039 | 0.1619 ± 0.0052 |

The results show that Regularized Time-Distributed MLP consistently shows low val_rmse and val_loss across all ratios, demonstrating excellent balance and generalization, suggesting that regularization in Time-Distributed MLP is particularly effective in reducing overfitting and maintaining generalization on validation data. Another observation is that regularization consistently improves validation performance by preventing overfitting. This is evident in the Regularized Time-Distributed MLP outperforming its non-regularized counterpart. The graphic results of dynamic RMSE and Loss values during the training and validation process at different ratios for dataset 1 are provided in the supplementary materials.

At higher training ratios, most models exhibit slightly increased validation RMSE, implying that a smaller training set might provide better generalization in this context. The train and validation RMSE values for the best-performing model are close in magnitude, suggesting that the model is robust and not significantly overfitting. It should also be noted that while models such as LSTM and CNN have demonstrated strong performance in other cases, they were not the best performers here. This might indicate that this specific dataset benefits more from Time Distributed MLP architectures. This could be due to the sequential or hierarchical nature of the data, which aligns well with the Time Distributed MLP's structure. Table 8 lists the best models for each ratio based on their val_rmse and val_loss values.

Table 8. Summary of RMSE Results

| Ratio | Best Model | RMSE |
| --- | --- | --- |
| 0.2 | Reg. Time Distributed MLP | 0.1213 |
| 0.3 | Time Distributed MLP | 0.1387 |
| 0.4 | Reg. Time Distributed MLP | 0.1442 |
| 0.5 | LSTM | 0.1698 |

At a ratio of 0.2, the model achieved both low train and validation RMSE, showing a good balance between training and generalization. At a ratio of 0.3, the validation RMSE increased compared to the train RMSE, which might indicate slight overfitting as the training ratio increased. Figures 6 and 7 illustrate the Regularized Time-Distributed MLP forecasting performance for lower ratios, where regularization produces better results. The figures are designed to use the region highlighted in blue as a control area for the forecast values. The figures clearly show that the predicted values of DL methods are within the control limits.

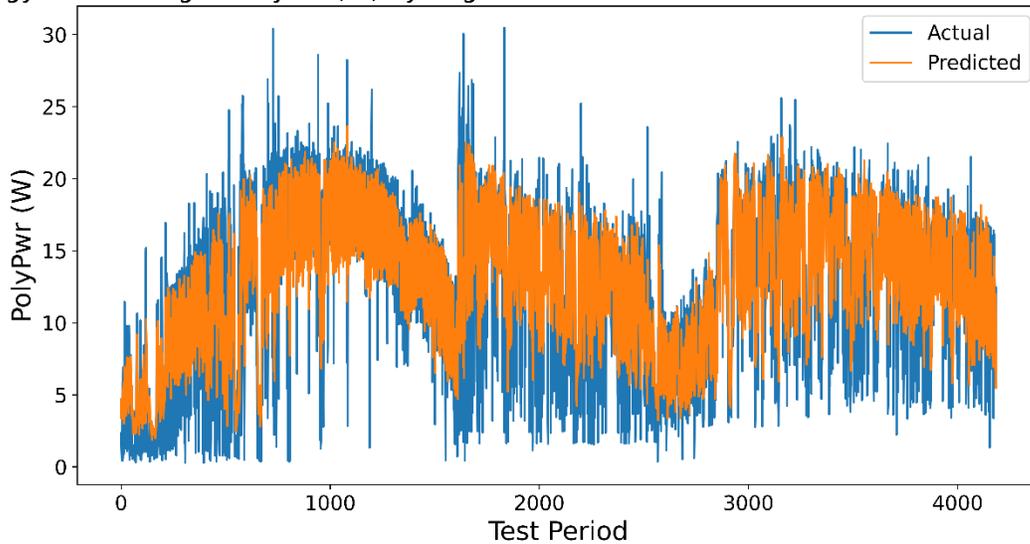

Figure 6. Regularized Time Distributed MLP at 20% for Test Data

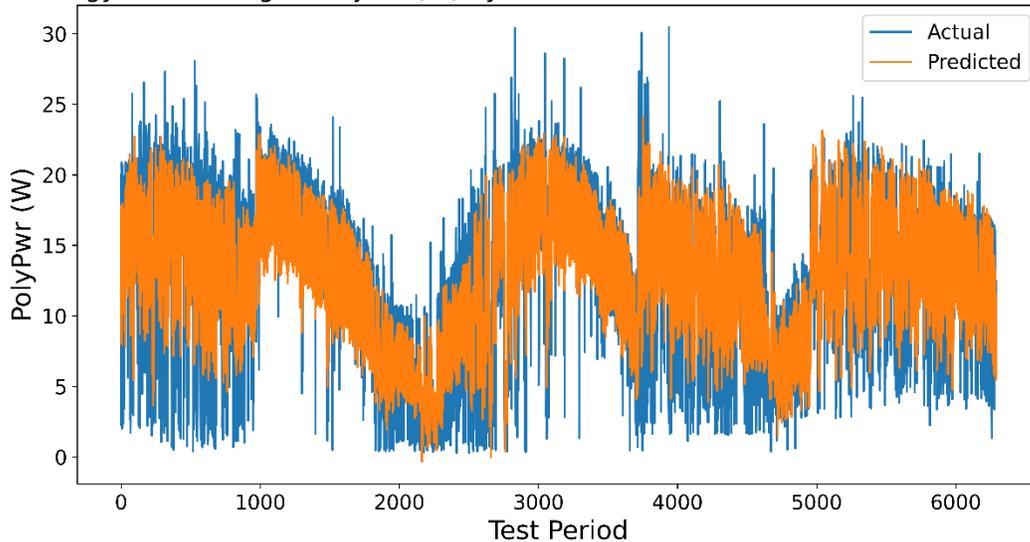

Figure 7. Time Distributed MLP at 30% for Test Data

Figures 8 and 9 illustrate the comparison of actual values versus the forecast of Time-Distributed MLP and LSTM for ratio = 0.4 and 0.5, respectively.

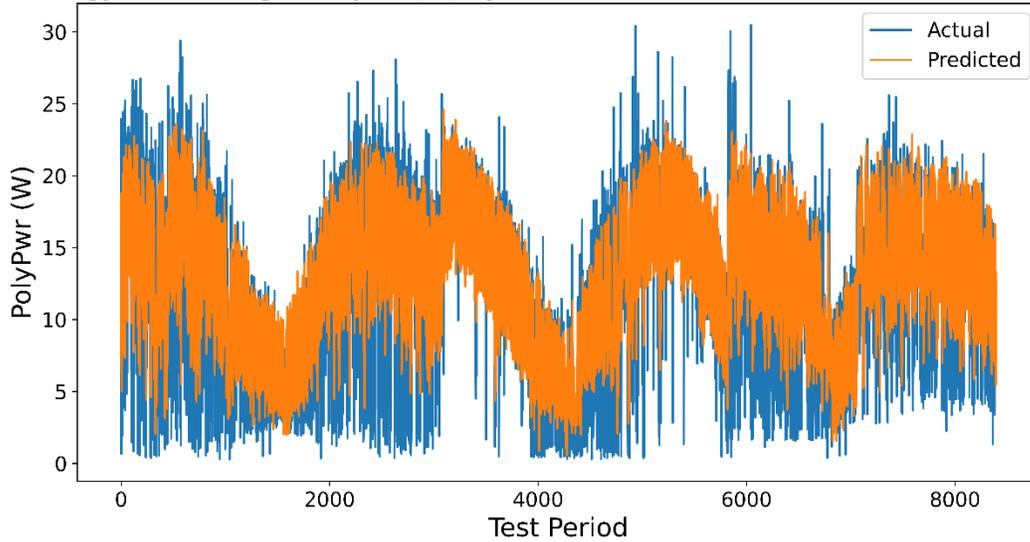

Figure 8. Time-Distributed MLP at 40% for Test Data

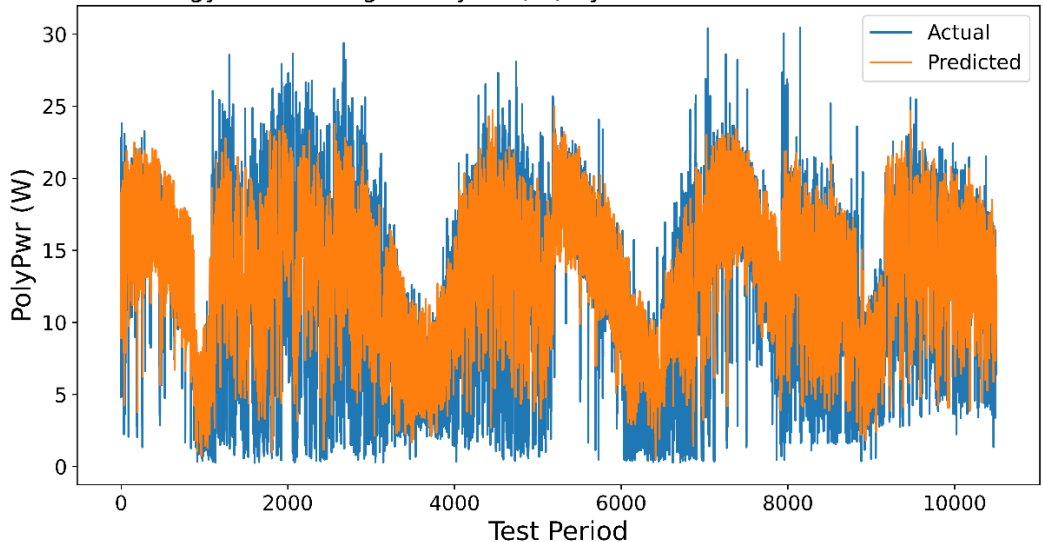

Figure 9. LSTM at 50% for Test Data

DL prediction models offer practical benefits for feasibility studies involving the renewable energy capacity of the systems. Hourly weather features are analyzed here to predict renewable energy generation. The LSTM model performs exceptionally well. The validation data exhibits a significant RMSE of approximately 0.04. Moreover, the loss function of the LSTM method results in superior results, as illustrated in the supplemental materials.

For the first dataset, CNN-LSTM is more stable than CNN alone, particularly performing well for smaller ratios, while the Encoder-Decoder method outperforms others for larger ratios. The results show that different ratios can make a difference in the performance of DL models. The Regularized Time Distributed MLP is the best-performing method in the second dataset, excelling in balancing training and validation performance, especially with smaller training data ratios. It highlights the importance of regularization and architectural choice in achieving optimal results.

LSTMs and Encoder-Decoders often require extensive datasets to learn dependencies over long sequences, while smaller training sizes lead to underfitting and increased error. Meanwhile, CNNs are relatively less sensitive to data size because convolutional layers efficiently capture hierarchical features (Simonyan and Zisserman, 2015). Stacked LSTM architectures excel with larger datasets but overfit quickly when the training size reduces (Sutskever et al., 2014).

While most models perform well with large training datasets (ratio = 0.2), CNN and CNN-LSTM particularly thrive owing to their ability to extract complex spatial and temporal features. Meanwhile, DNN and simpler architectures may be overfitting if not regularized. As the ratio increases, temporal models such as LSTM, CNN-LSTM, and Encoder-Decoder struggle as they require more data to capture dependencies. DNN and Time-Distributed MLP may overfit or fail to generalize due to insufficient training examples. On the other hand, CNN is still effective owing to its reliance on feature hierarchies.

The observed variations in model performance across the two datasets underscore the importance of aligning model architecture with dataset characteristics. Dataset-1, comprising multi-year hourly weather and power generation data, is more temporally rich and exhibits pronounced seasonality and long-term dependencies. Consequently, temporal models such as LSTM, Stacked LSTM, and Encoder-Decoder performed better due to their strength in modeling sequential patterns. In contrast, Dataset-2 contains photovoltaic power outputs influenced more heavily by instantaneous weather conditions and spatial attributes, with weaker temporal continuity. Here, models like the Time-Distributed MLP and standard MLP showed superior performance, likely because they are better suited for learning complex nonlinear relationships among static or lightly-sequenced variables. These differences highlight the need to assess data structure—such as temporal density, sequence relevance, and spatial variance—when selecting or designing deep learning models for renewable energy forecasting. Table A5 provides a summary of model suitability based on dataset characteristics.

While the study demonstrates strong predictive performance across two datasets with diverse temporal and spatial characteristics, it is important to acknowledge the limitation of lacking external validation on independent datasets beyond those analyzed. As a result, the generalizability of the proposed models to unseen geographies, weather patterns, or energy systems remains to be empirically verified. Future research should incorporate cross-regional or cross-technology validation using publicly available benchmarks or real-time operational data from distinct renewable infrastructures. Such external validation would enhance the robustness of the findings and confirm the adaptability of the proposed framework across varied energy forecasting scenarios. Furthermore, model transferability through techniques such as transfer learning or domain adaptation could be explored to extend the application of trained models to similar but not identical contexts.

### 4.3 Statistical Significance Analysis

To determine whether the performance differences among multiple models are statistically significant, Friedman test is applied (Friedman, 1937). It is a non-parametric test that ranks the models on each dataset instead of using raw performance scores. This approach allows for the comparison of multiple models across multiple datasets, where each dataset corresponds to a different training and validation split ratio. In addition, Friedman's test does not assume normality of the performance metric, making it suitable for a wide range of model evaluation scenarios (Demsar, 2006). The results of the test are presented in Table 9.

Table 9. Friedman's test results

|  |  | Dataset-1 | | Dataset-2 | |
| --- | --- | --- | --- | --- | --- |
| Evaluation Metrics | df | chi-squared | p-value | chi-squared | p-value |
| Train RMSE | 14 | 53.3828 | 1.64E-06 | 50.0285 | 6.04E-06 |
| Validation RMSE | 14 | 49.6380 | 7.02E-06 | 32.8572 | 0.003021 |
| Train MAE | 14 | 53.3789 | 1.64E-06 | 52.0339 | 2.78E-06 |
| Validation MAE | 14 | 49.9024 | 6.34E-06 | 41.5805 | 0.000144 |
| Train R-Square | 14 | 53.1500 | 1.79E-06 | 49.4000 | 7.69E-06 |
| Validation R-Square | 14 | 50.3250 | 5.39E-06 | 33.0084 | 0.002873 |
| Train Loss | 14 | 53.5205 | 1.55E-06 | 50.2647 | 5.51E-06 |
| Validation Loss | 14 | 53.0722 | 1.85E-06 | 39.9097 | 0.000264 |

It can be observed the p-values are below 0.05 for all evaluation metrics, indicating the null hypothesis, which assumes all models have equal performance, is rejected. Thus, it can be concluded that statistically significant differences exist in performance of the models across all metrics for training and validation data.

### *4.4 Computation Feasibility for Deployment*

Although this research evaluates the predictive accuracy of DL methods in forecasting renewable energy, taking their computational requirements into account is also important for practical applications. DL models such as CNN-LSTM, Stacked LSTM, and Encoder-Decoder require higher computation resources due to their depth and complexity, resulting in longer training times and higher memory usage.

To contextualize resource requirements, preliminary benchmarks show that simpler models such as DNN and Time-Distributed MLP can train in 10 to 30 minutes on a mid-range GPU, whereas deeper models such as Stacked-LSTM and Encoder-Decoder may demand more than an hour under the same conditions. Inference times also show variance, with MLP and CNN architectures being more appropriate for real-time forecasting owing to their forward-pass computation. Finally, the number of trainable parameters effect the optimization process as well. Simpler models contain less than 500 thousand parameters while that number for the Stacked LSTM and Encoder-Decoder models can reach more than a million.

Such computation variations indicate the significance of balancing efficiency with predictive performance, especially in operational settings where real-time requirements are critical. To this end, it is important for the future studies to include standardized benchmarks for training time, parameter count, and inference latency to inform model selection for energy forecasting systems. Table A6 provides the parameter counts for each model based on the configurations. While these approximations are dependent on the exact input/output shapes, they clearly indicate that LSTM-based and hybrid models are significantly more resource-intensive compared to simple MLP or CNNs.

### 5. Future Research Directions

While the proposed framework demonstrates high predictive accuracy across multiple DL architetures and datasets, several limitations exist. The complexity of model development, including preprocessing steps such as hyperparameter tuning, PCA, mutual information filtering, and stationarity testing, can increase computational cost. Although regularization techniques mitigate overfitting, fine-tuning these parameters requires significant domain expertise and iterative testing. Moreover, the performance of the proposed framework depends on the completeness and quality of input data. Missing data, sensor errors, or abrupt changes in weather conditions may reduce prediction accuracy. Another limitation relates to the interpretability of DL models; despite their high accuracy, black-box behavior can hinder their adoption in energy management decisions. Finally, while the training-test ratio sensitivity analysis improves model robustness, the approach assumes stationarity and representative distribution, which may not hold in all future forecasting environments.

While the proposed models demonstrate high average accuracy across two datasets, their performance under extreme conditions such as sudden demand spikes or rare weather events remains unexplored. This research covers time-series characteristics like stationarity and seasonality, regularization to handle overfitting, and model performance under different train/test ratios. Extreme cases can significantly impact grid reliability and operational planning. Deep learning models often underperform on such outliers due to the lack of sufficient anomalous samples during training. Future work should incorporate scenario-based testing or anomaly-aware validation to assess model resilience. One approach is to simulate synthetic weather shocks or demand surges and evaluate prediction drift under these scenarios. Additionally, techniques like adversarial training, uncertainty quantification, or hybrid models with rule-based overrides can improve robustness in operational settings.

## 5  Conclusion

One of the main contributions of this study is the application of many ML methods comparatively to a set of applications in renewable energy areas. DL techniques were tested on two datasets with four different training/test ratios, showing their relative performance. The study responds to the need for a robust DL method that can be utilized in many applications related to renewable energy. The prediction models employed in the study performed exceptionally well when the climate factors were used as predictors.

Traditional LSTM networks showed consistent results, particularly in handling long-term dependencies. The stacked variant further enhanced performance by leveraging deeper architectures, albeit at the cost of increased computational complexity. CNN-LSTM and Encoder-Decoder models exhibited superior performance in capturing spatial and temporal features, making them particularly well-suited for complex time-series data. Their robustness across varying training ratios highlights their potential for real-world applications with inconsistent data availability. CNNs and Time-Distributed MLP methods demonstrated competitive performance in cases where spatial correlations dominated. On the other hand, their effectiveness diminished when required to model complex temporal patterns. While DNNs provided baseline performance, their inability to effectively capture sequential dependencies limited their utility for time-series forecasting tasks.

The findings underscore the necessity of model selection customized for the particular characteristics of the dataset and the prediction task. The superior performance of hybrid architectures like CNN-LSTM and Encoder-Decoder models suggests that integrating spatial and temporal feature extraction is critical for advancing forecasting accuracy in renewable energy. The comparative analysis under different training/test ratios revealed the importance of data sufficiency for model generalizability. The larger training data size generally improved performance, underscoring the value of comprehensive data collection in renewable energy projects.

**Data Availability Statement**

A complete set of figures generated by the methods during and/or analyzed during the current study is available in the figshare repository, https://doi.org/10.6084/m9.figshare.28075349.

**Declaration of Generative AI**

During the preparation of this work, the authors used Grammarly to improve the readability and language of the manuscript. After using this service, the authors reviewed and edited the content as needed and take full responsibility for the content of the published article.

# Appendix

Table A1. ADF Stationary Test Results for Features of Dataset 1 Used in DL models

| Variable Code | Variables | ADF Statistic | KPSS Statistic | p-value | Lags Used |
|---|---|---|---|---|---|
| 1 | generation_hydro | -10.8656 | 1.6317 | 1.41E-19 | 49 |
| 2 | generation_other_renewable | -7.9666 | 26.3764 | 2.85E-12 | 50 |
| 3 | generation_solar | -13.8797 | 0.8648 | 6.24E-26 | 52 |
| 4 | generation_wind | -17.8878 | 0.2054 | 2.99E-30 | 49 |
| 5 | renewable_total | -14.9115 | 0.8063 | 1.46E-27 | 49 |
| 6 | temp_Barcelona | -5.0137 | 0.5734 | 2.08E-05 | 51 |
| 7 | temp_min_Barcelona | -5.8899 | 0.9342 | 2.94E-07 | 52 |
| 8 | temp_max_Barcelona | -4.4326 | 0.5139 | 0.00026 | 50 |
| 9 | pressure_Barcelona | -13.6666 | 0.4352 | 1.49E-25 | 52 |
| 10 | humidity_Barcelona | -13.9506 | 0.7225 | 4.71E-26 | 52 |
| 11 | wind_speed_Barcelona | -15.7979 | 7.5828 | 1.08E-28 | 51 |
| 12 | wind_deg_Barcelona | -15.0506 | 3.5863 | 9.31E-28 | 48 |
| 13 | rain_1h_Barcelona | -12.3064 | 1.2461 | 7.26E-23 | 52 |
| 14 | clouds_all_Barcelona | -23.7869 | 0.4019 | 0 | 28 |
| 15 | temp_Bilbao | -7.5654 | 0.6454 | 2.94E-11 | 52 |
| 16 | temp_min_Bilbao | -7.5656 | 0.4945 | 2.93E-11 | 52 |
| 17 | temp_max_Bilbao | -8.09679 | 1.1319 | 1.33E-12 | 49 |
| 18 | pressure_Bilbao | -12.3602 | 0.4994 | 5.57E-23 | 52 |
| 19 | humidity_Bilbao | -17.2311 | 0.2284 | 6.19E-30 | 52 |
| 20 | wind_speed_Bilbao | -17.951 | 1.6327 | 2.84E-30 | 52 |
| 21 | wind_deg_Bilbao | -18.386 | 1.8015 | 2.20E-30 | 52 |
| 22 | rain_1h_Bilbao | -14.9749 | 4.1601 | 1.19E-27 | 52 |
| 23 | snow_3h_Bilbao | -15.2003 | 0.5781 | 5.84E-28 | 52 |
| 24 | clouds_all_Bilbao | -18.6754 | 0.3816 | 2.04E-30 | 47 |
| 25 | temp_Madrid | -5.01856 | 0.5539 | 2.04E-05 | 52 |
| 26 | temp_min_Madrid | -5.3926 | 0.5183 | 3.51E-06 | 52 |
| 27 | temp_max_Madrid | -5.14762 | 0.5597 | 1.12E-05 | 49 |
| 28 | pressure_Madrid | -8.47753 | 4.6889 | 1.42E-13 | 52 |
| 29 | humidity_Madrid | -8.67858 | 0.4826 | 4.35E-14 | 51 |
| 30 | wind_speed_Madrid | -16.5852 | 0.6097 | 1.82E-29 | 52 |
| 31 | wind_deg_Madrid | -17.0531 | 1.9762 | 8.04E-30 | 52 |
| 32 | rain_1h_Madrid | -17.5915 | 0.1728 | 3.96E-30 | 47 |
| 33 | snow_3h_Madrid | -187.251 | 0.1252 | 0 | 0 |
| 34 | clouds_all_Madrid | -15.1352 | 0.2860 | 7.14E-28 | 45 |
| 35 | temp_Seville | -5.07516 | 1.3519 | 1.57E-05 | 52 |
| 36 | temp_min_Seville | -5.26475 | 0.5948 | 6.48E-06 | 52 |
| 37 | temp_max_Seville | -5.38049 | 3.1623 | 3.72E-06 | 51 |
| 38 | pressure_Seville | -12.4495 | 2.1130 | 3.60E-23 | 52 |
| 39 | humidity_Seville | -11.1226 | 0.5343 | 3.44E-20 | 52 |
| 40 | wind_speed_Seville | -19.6939 | 0.3487 | 0 | 52 |
| 41 | wind_deg_Seville | -14.9972 | 0.5218 | 1.10E-27 | 52 |
| 42 | rain_1h_Seville | -18.7032 | 0.3114 | 2.04E-30 | 50 |
| 43 | clouds_all_Seville | -15.5061 | 0.5598 | 2.38E-28 | 50 |
| 44 | temp_Valencia | -5.54853 | 1.1249 | 1.64E-06 | 52 |
| 45 | temp_min_Valencia | -5.72421 | 2.0593 | 6.83E-07 | 52 |
| 46 | temp_max_Valencia | -5.41683 | 0.6673 | 3.12E-06 | 52 |
| 47 | pressure_Valencia | -11.6734 | 0.7299 | 1.81E-21 | 51 |
| 48 | humidity_Valencia | -15.1886 | 0.3687 | 6.06E-28 | 52 |
| 49 | wind_speed_Valencia | -15.6497 | 0.5155 | 1.60E-28 | 52 |
| 50 | wind_deg_Valencia | -14.0857 | 0.9936 | 2.78E-26 | 52 |
| 51 | rain_1h_Valencia | -14.4376 | 1.3168 | 7.45E-27 | 48 |

| 52 | snow_3h_Valencia | -24.4656 | 0.3287 | 0 | 52 |
| 53 | clouds_all_Valencia | -17.8195 | 0.3121 | 3.17E-30 | 46 |

Table A2. ADF Stationary Test Results for Features of Dataset 2 Used in DL models

| Variable Code | Variables | ADF Statistic | KPSS | p-value | Lags Used |
|---|---|---|---|---|---|
| 1 | AmbientTemp | -6.51046 | 0.33894 | 1.10E-08 | 46 |
| 2 | Cloud.Ceiling | -15.3386 | 3.02255 | 3.86E-28 | 38 |
| 3 | cos_hr | -16.3438 | 0.48979 | 2.98E-29 | 41 |
| 4 | cos_mon | -6.05791 | 0.12435 | 1.23E-07 | 0 |
| 5 | Humidity | -9.0394 | 2.21972 | 5.18E-15 | 46 |
| 6 | Latitude | -3.03967 | 1.04945 | 0.031352 | 0 |
| 7 | PolyPwr | -8.45448 | 0.45103 | 1.63E-13 | 46 |
| 8 | Season_Spring | -6.30732 | 0.13124 | 3.30E-08 | 0 |
| 9 | Season_Summer | -5.98208 | 0.22521 | 1.83E-07 | 0 |
| 10 | Season_Winter | -6.26504 | 0.19708 | 4.13E-08 | 0 |
| 11 | sine_hr | -15.8986 | 0.35433 | 8.38E-29 | 46 |
| 12 | sine_mon | -4.09225 | 0.39871 | 0.000997 | 0 |
| 13 | Visibility | -22.4918 | 0.42861 | 0 | 24 |
| 14 | Wind.Speed | -10.9548 | 1.97364 | 8.62E-20 | 46 |

Table A3. Predictive models

| Long-Short Term Memory (LSTM): |
|---|
| This approach uses recursive connections instead of linear and sequential data processing, allowing it to identify intricate patterns and relationships in data sequences. It can obtain deeper insights and increase accuracy in sequence processing jobs by considering the complete data sequence rather than individual data points separately. Because of this feature, LSTM is now widely used in various applications, such as time series forecasting and natural language processing (Kong et al., 2019). |
| Stacked LSTM: |
| The model is a DL technique that can provide superior performance on various prediction applications by learning increasingly abstract representations of the input data. It is an extension of the conventional LSTM model, which consists of a single hidden layer of LSTM cells. The depth and complexity of the network are increased by this new architecture's numerous hidden layers with multiple memory cells (Yu et al., 2019). One important component of this model is its depth, which enables it to identify intricate relationships and patterns in the data that shallower models can miss (Atef al., 2020). |
| Convolutional Neural Networks (CNN): |
| CNN is a DL model architecture best suited for handling grid-like input, like images. This model can automatically find and depict linkages and intricate patterns in the data by learning spatial hierarchies of features in a self-organizing way (Abdoos et al., 2024). Because of this capability, CNN has become a useful tool for various computer vision applications, such as segmentation, image recognition, and object detection. |
| CNN-LSTM: |
| A combined method was developed in this study, utilizing the strengths of CNN and LSTM to achieve improved classification. The CNN extracts intricate features from the data, while the LSTM network serves as a classifier, utilizing its internal memory to learn from past experiences and capture long-term dependencies. The LSTM network is designed to process sequential data in a specific order, allowing it to capture temporal relationships and patterns, whereas traditional fully connected networks process only one input at a time (Agga et al., 2022). Combining these features significantly improves classification accuracy, making this a promising solution for various applications (Ku et al., 2020). |

DNN:

This is a neural network designed to learn complex patterns and relationships in data. Unlike traditional NN, which have only a few layers, DNNs have many layers that work together to process and transform the input data. The output of each layer is used as input to the next one, allowing the network to learn abstract and complex representations of the data. The multilayer architecture of DNNs enables them to perform complex computational tasks, such as executing multiple complex operations simultaneously, making them particularly appropriate for tasks requiring intense computational power and complex data analysis (Oluleye et al., 2023).

The Multilayer Perceptron (MLP):

MLP is a NN architecture with a hierarchical structure, comprising an input layer, multiple hidden layers, and an output layer. Each layer comprises a collection of neurons, which are the fundamental components of the network. These neurons work together to process and transform the input data, enabling the MLP to learn and represent complex patterns and relationships in the data. The MLP's multilayer architecture allows it to capture and model complex interactions between variables, making it a powerful tool for various applications (Chan et al., 2023).

Encoder-Decoder:

This framework is a type of NN employing a multi-scale feature extraction approach, using dense convolutional and transition layers to capture features at various scales. The decoder part of the network plays a crucial role in recovering lost information. It is connected to the encoder part through long-range skip connections, which allow the network to propagate information efficiently between the encoder and decoder, enabling the network to learn complex patterns and relationships in the data and speed up the training process (Wang et al., 2022).

Table A4. Final Hyperparameter Configurations

| Model | Layers | Neurons per Layer | Activation Function | Learning Rate | Dropout Rate | Batch Size | Optimizer |
|---|---|---|---|---|---|---|---|
| LSTM | 3 | 64, 64, 32 | tanh | 0.001 | 0.3 | 64 | Adam |
| Stacked LSTM | 4 | 128, 128, 64, 32 | tanh | 0.001 | 0.3 | 64 | Adam |
| CNN | 2 conv + 1 dense | 64, 128 + 64 | ReLU | 0.0005 | 0.25 | 32 | Adam |
| CNN-LSTM | 2 conv + 2 LSTM | 64, 128 (CNN), 64 (LSTM) | ReLU / tanh | 0.001 | 0.3 | 64 | Adam |
| Deep Neural Network | 4 | 256, 128, 64, 32 | ReLU | 0.001 | 0.4 | 128 | Adam |
| Time-Distributed MLP | 3 | 128, 64, 32 | ReLU | 0.0005 | 0.2 | 64 | RMSprop |
| Encoder-Decoder | 2 enc + 2 dec | 128 (enc), 64 (dec) | Tanh | 0.0005 | 0.3 | 64 | Adam |

Table A5. Model Suitability Based on Dataset Characteristics

| Model | Best Suited For | Characteristics |
|---|---|---|
| LSTM / Stacked LSTM | Datasets with strong temporal dependencies | Captures long-term trends, seasonality, and autocorrelation in time series |
| Encoder-Decoder | Multi-step sequence prediction, longer sequences | Learns variable-length input/output sequences; effective for structured temporal data |
| CNN-LSTM | Datasets with both spatial and temporal patterns | Extracts spatial features (e.g., weather sensors), then models sequential dependencies |
| CNN | Grid-like or spatially structured input data | Identifies spatial hierarchies but limited for modeling temporal continuity |
| MLP / Time-Distributed MLP | Datasets with static or weak temporal structure | Efficiently models nonlinear relationships among features; suitable for tabular weather data |

| DNN | High-dimensional static data | Handles complex nonlinear mappings but may overfit without temporal cues |

Table A6. Estimated Parameter Count Summary

| Model | Layers / Architecture | Parameter counts | Notes |
|---|---|---|---|
| LSTM | 3 layers (64, 64, 32) | ~100K–150K | Includes recurrence |
| Stacked LSTM | 4 layers (128, 128, 64, 32) | ~400K–600K | Higher due to deep memory |
| CNN | 2 conv (64, 128) + dense (64) | ~50K–150K | Lightest among complex |
| CNN-LSTM | 2 conv + 2 LSTM (64, 128 CNN + 64 LSTM) | ~300K–500K | High due to dual feature extraction |
| DNN | 4 layers (256, 128, 64, 32) | ~300K–450K | Fully connected, no recurrence |
| Time-Distributed MLP | 3 layers (128, 64, 32) | ~150K–250K | Efficient for spatial learning |
| Encoder-Decoder | 2 enc (128), 2 dec (64) | ~400K–600K | Heavy due to skip connections |

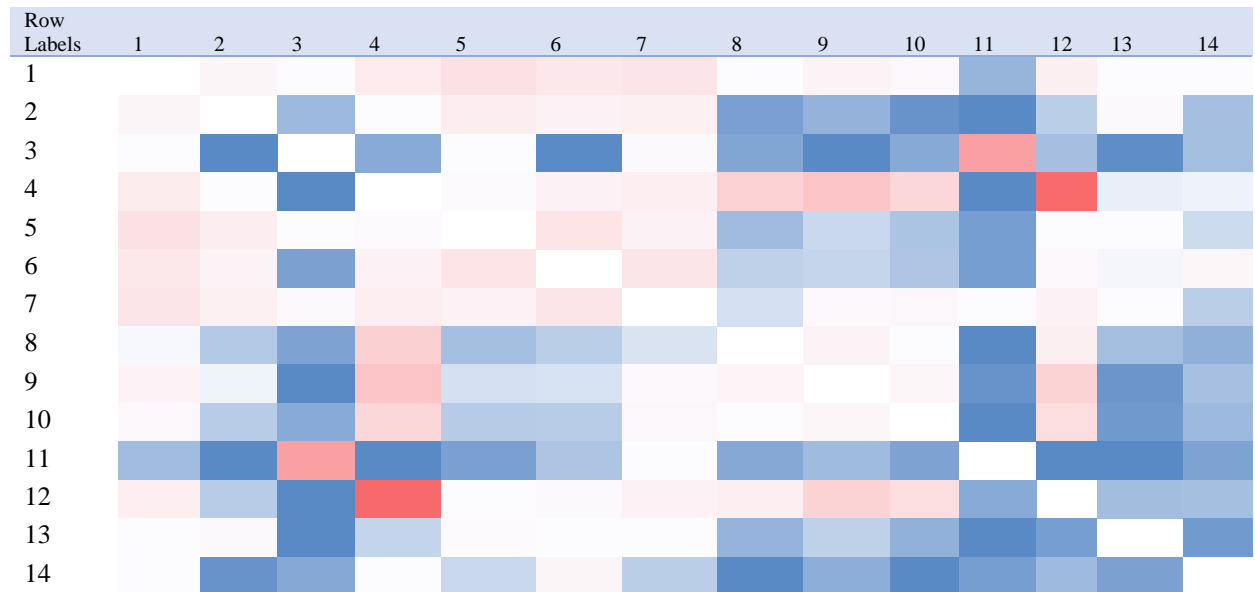

Figure A1. Mutual Information Dataset-2

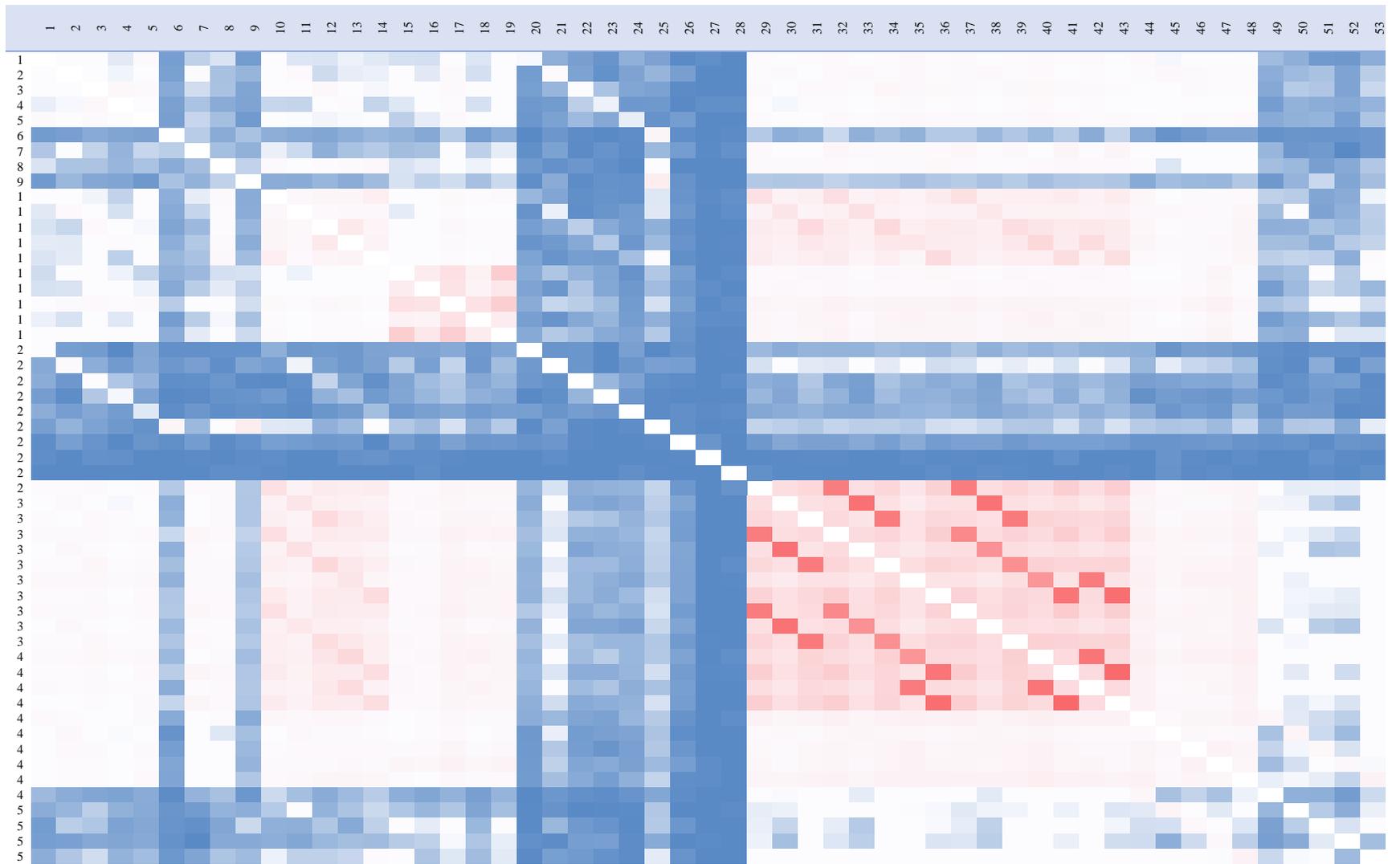

Figure A2. Mutual Information Dataset-1

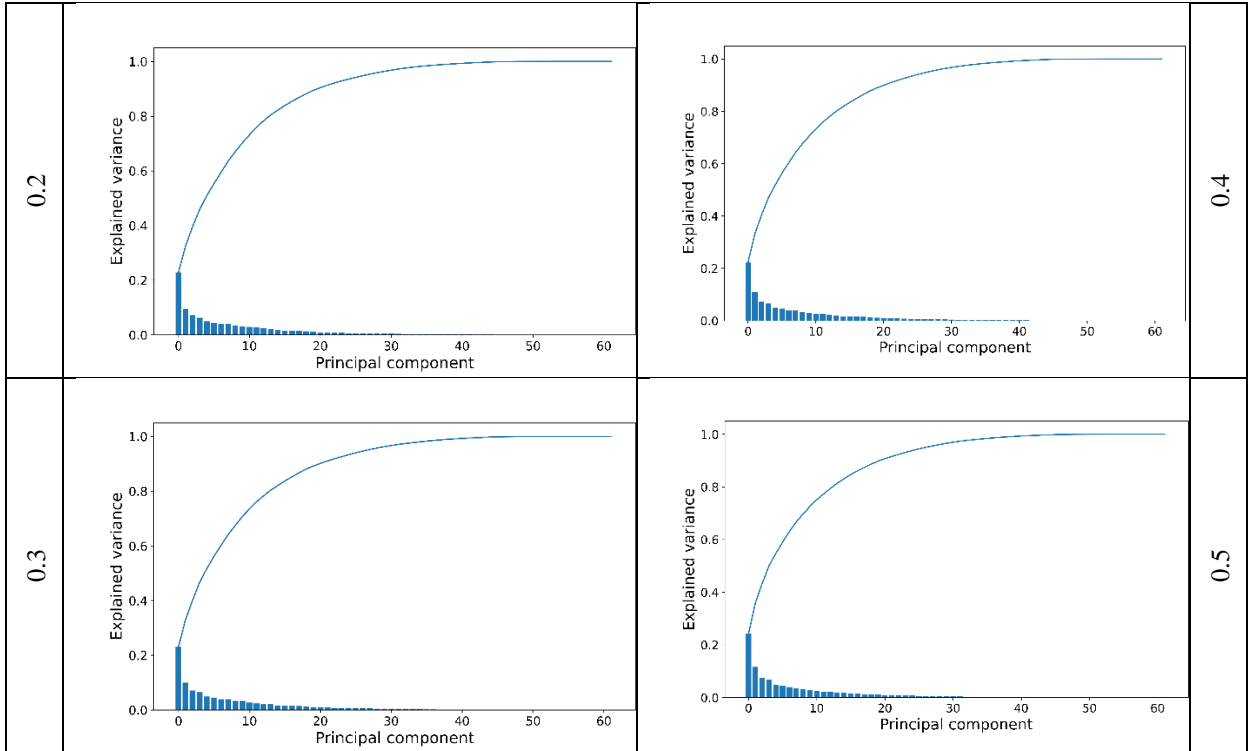
Figure A3. PCA Results for Dataset-1

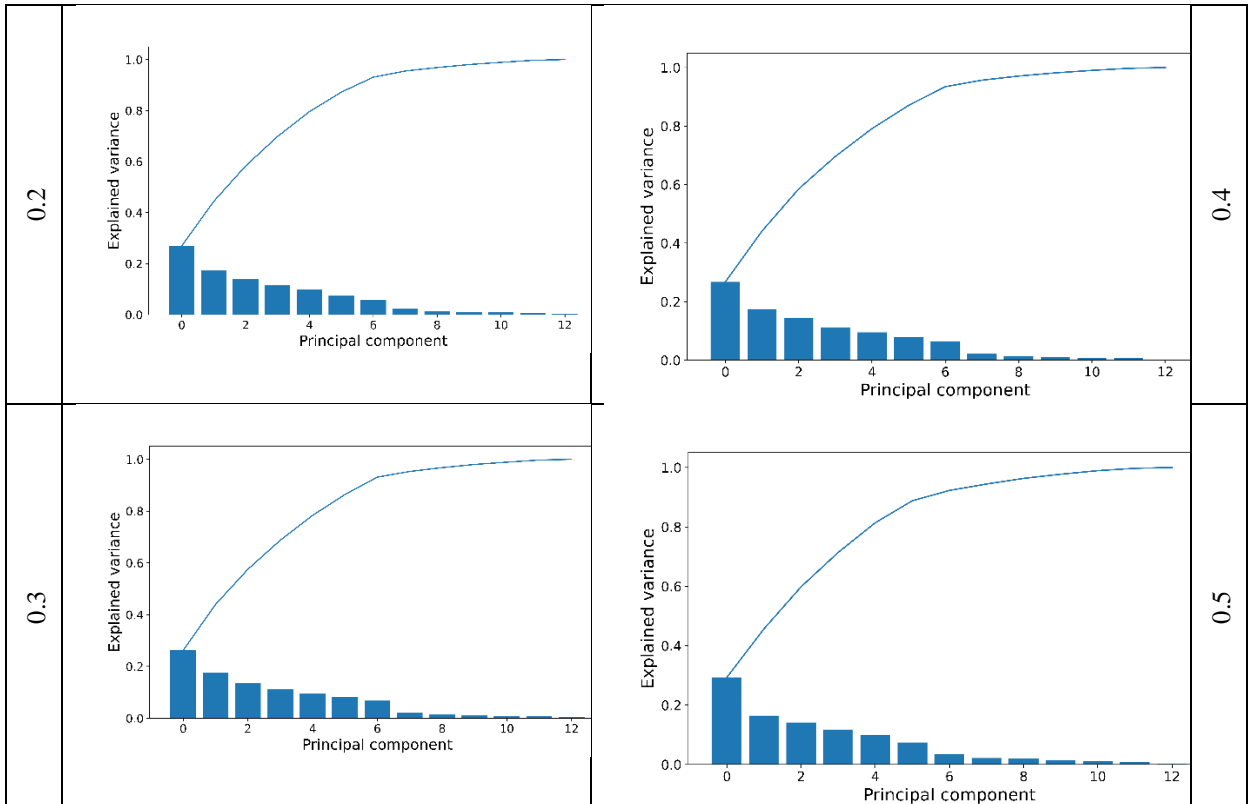
Figure A4. PCA Results for Dataset-2

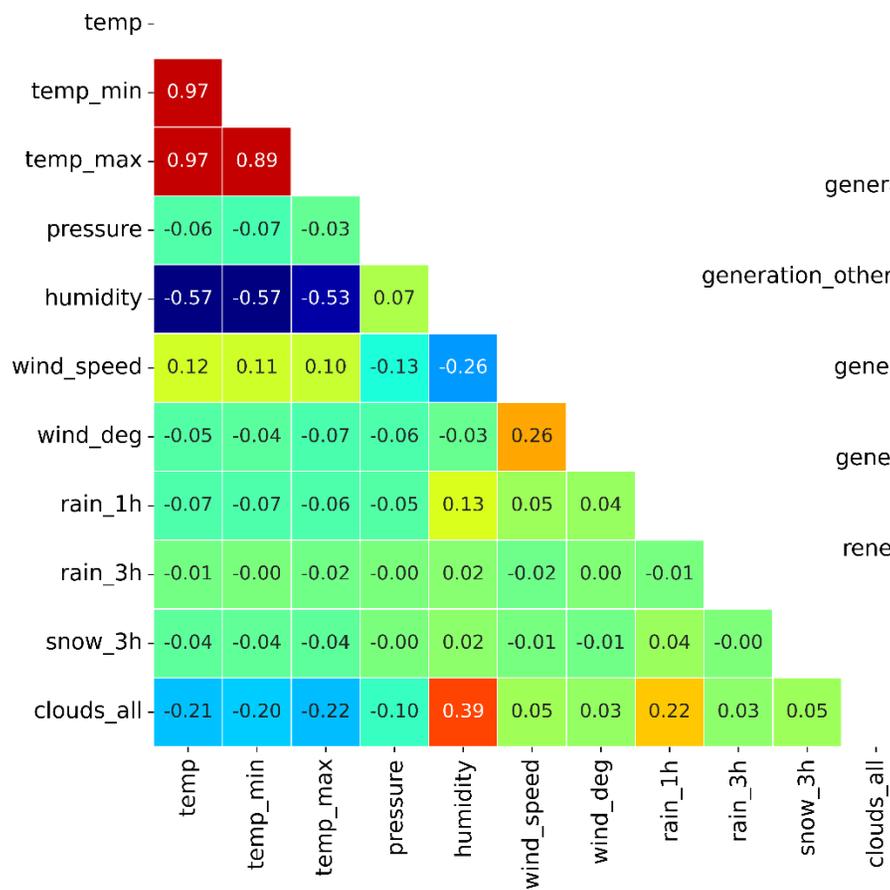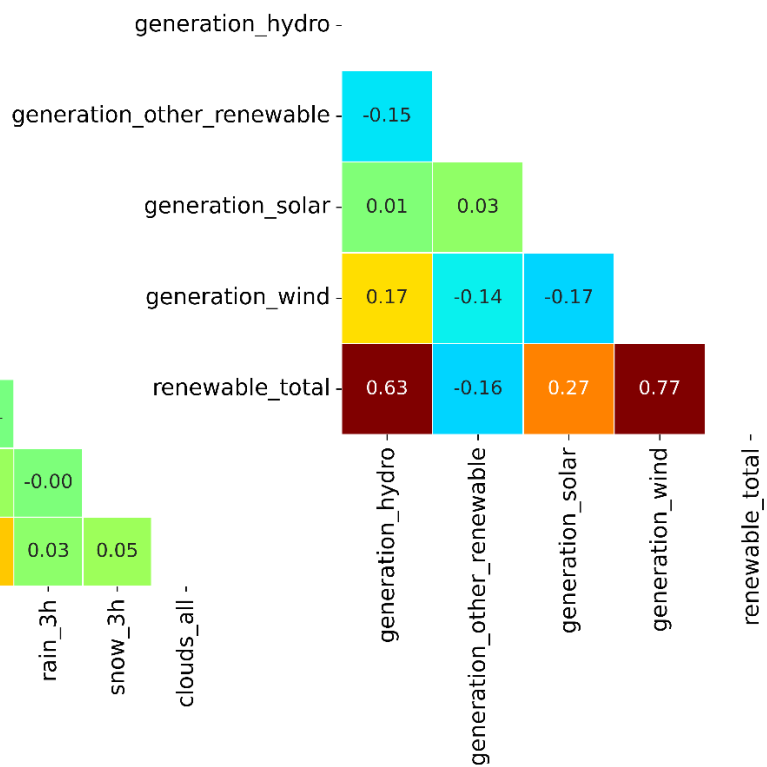

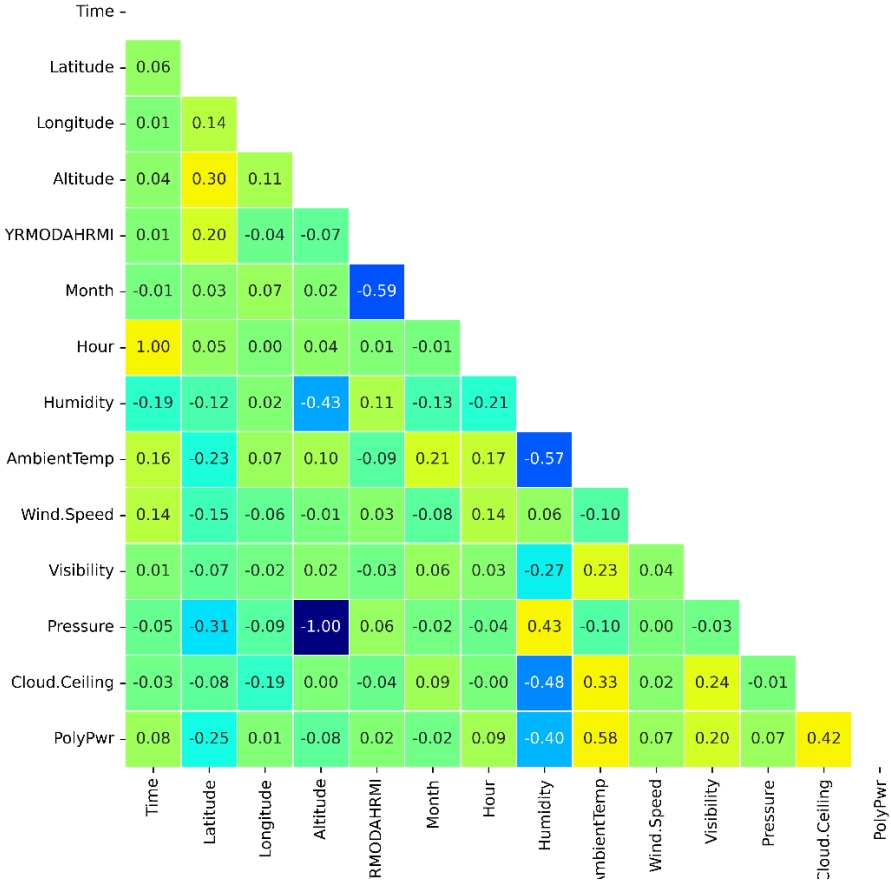

Figure A5. Correlation Analysis of Dataset 1

Figure A6. Correlation Analysis of Dataset 2

---

i https://greenfin.it/events/hourly-energy-demand-generation-and-weather
ii https://data.mendeley.com/datasets/hfhwmn8w24/5

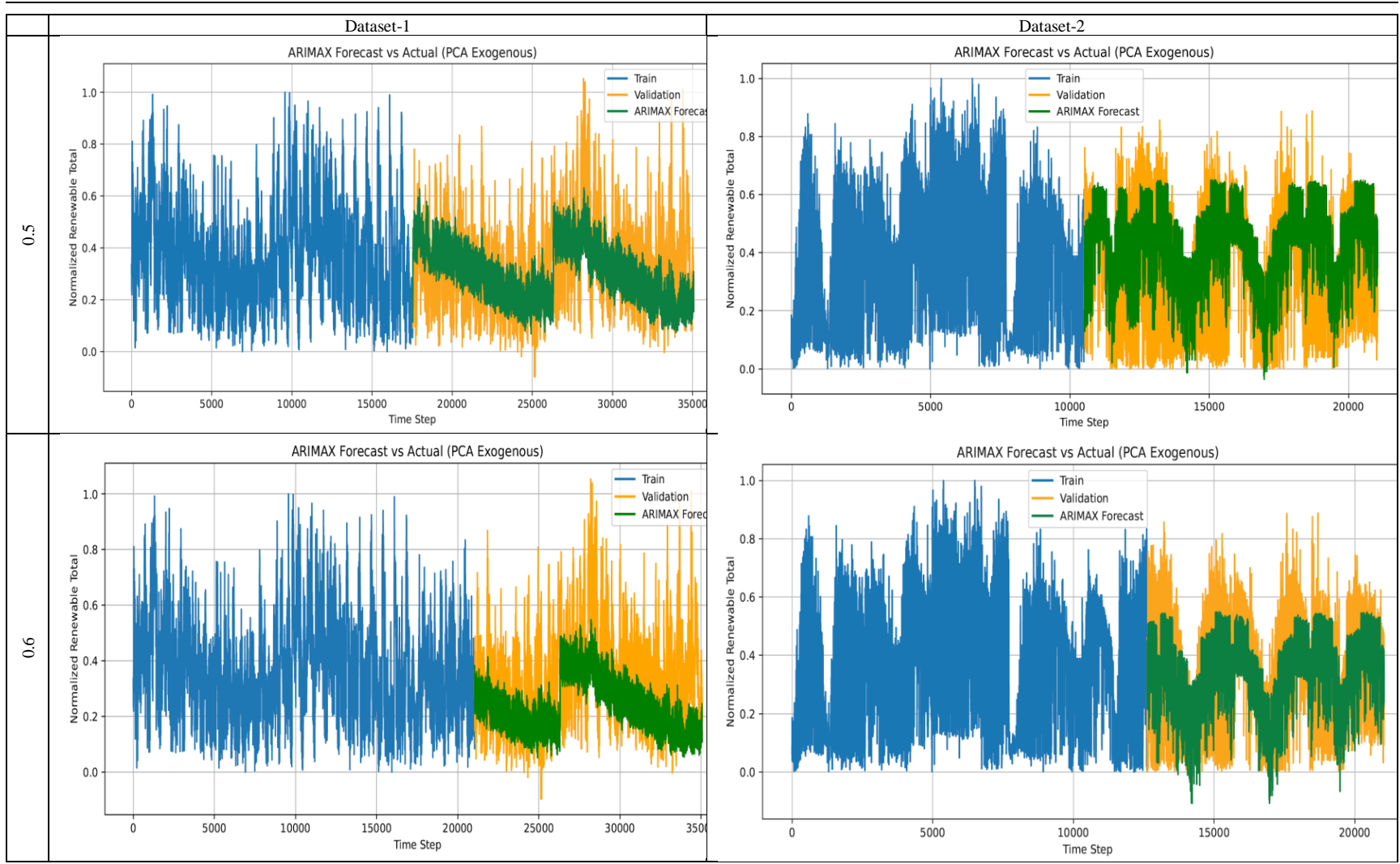

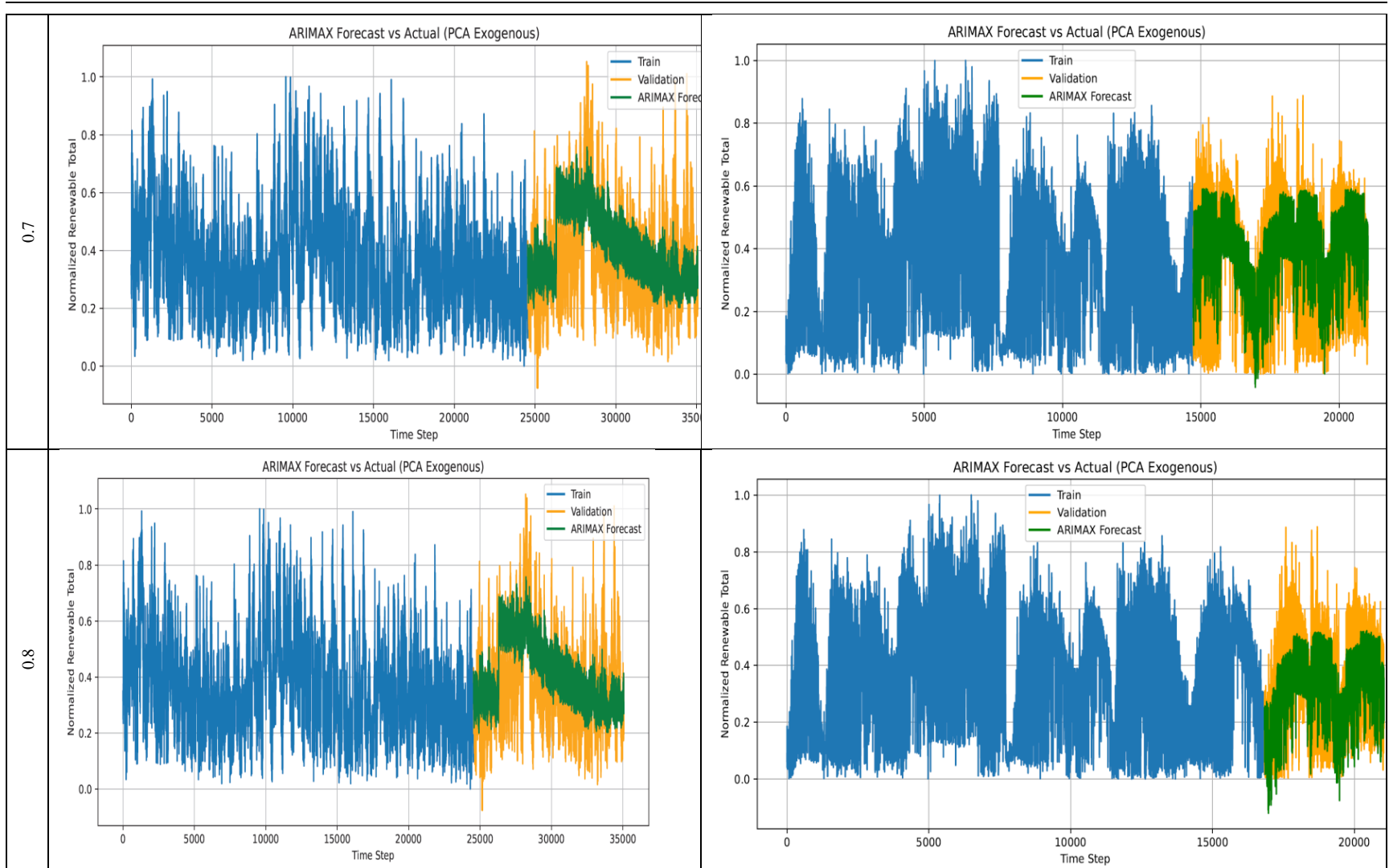

Figure A7. ARIMA Results